\documentclass{article}

\usepackage{amsmath,amsfonts,bm}

\def\eqref#1{equation~\ref{#1}}

\def\1{\bm{1}}

\DeclareMathAlphabet{\mathsfit}{\encodingdefault}{\sfdefault}{m}{sl}
\SetMathAlphabet{\mathsfit}{bold}{\encodingdefault}{\sfdefault}{bx}{n}

\usepackage{graphicx}
\usepackage{overpic}
\usepackage{booktabs, multirow, makecell} 
\usepackage{float}
\usepackage{hyperref}
\hypersetup{
  colorlinks=true,
  linkcolor=black,
  urlcolor=blue
}
\usepackage{url}
\usepackage[capitalize]{cleveref}
\crefname{section}{Sec.}{Secs.}
\usepackage[ampersand]{easylist}

\usepackage{adjustbox}
\usepackage{wrapfig}
\usepackage{amsmath}
\usepackage{amssymb}
\usepackage{dsfont}
\usepackage{colortbl}
\usepackage{subcaption}
\usepackage{xcolor}

\newcommand{\method}{GIM}
\newcommand{\methodlong}{the Generalized Induction-Head Model}
\newcommand{\fuzzyslmlong}{Fuzzy Matching Model}
\newcommand{\infinigram}{Infini-gram}

\newcommand{\exactinduction}{GIM (exact)}
\newcommand{\fuzzyinduction}{GIM (fuzzy)}

\newcommand{\err}[1]{\scriptsize $\pm${#1}}

\newcommand{\new}[1]{#1}

\usepackage{placeins}
\usepackage{xcolor}
\usepackage{marginnote}

\PassOptionsToPackage{numbers, compress}{natbib}

    \usepackage[final]{neurips_2025}

\usepackage[utf8]{inputenc} 
\usepackage[T1]{fontenc}    
\usepackage{hyperref}       
\usepackage{url}            
\usepackage{booktabs}       
\usepackage{amsfonts}       
\usepackage{nicefrac}       
\usepackage{microtype}      
\usepackage{xcolor}         

\title{Interpretable Next-token Prediction\\via the Generalized Induction Head}

\author{
\And
Eunji Kim$^{1, 2}$\thanks{Equal contribution. Work conducted during an internship at Microsoft Research.}
\And
Sriya Mantena$^{1, 3}$\footnotemark[1]
\And
\AND
Weiwei Yang$^1$
\And
Chandan Singh$^1$
\And
Sungroh Yoon$^{2,4}$\thanks{Corresponding authors.}
\And
Jianfeng Gao$^{1}$\footnotemark[2]
\AND\And\\
$^1$ Microsoft Research \\
$^2$ Department of Electrical and Computer Engineering, Seoul National University\\
$^3$ Stanford University\\
$^4$ Interdisciplinary Program in Artificial Intelligence, Seoul National University\\
}

\begin{document}

\maketitle

\begin{abstract}
While large transformer models excel in predictive performance, their lack of interpretability restricts their usefulness in high-stakes domains.
To remedy this, we propose \methodlong{} (\method{}),
an interpretable model for next-token prediction inspired by the observation of ``induction heads'' in LLMs.
\method{} is a retrieval-based module that identifies similar sequences in the input context by combining exact n-gram matching and fuzzy matching based on a neural similarity metric.
We evaluate \method{} in two settings: language modeling and fMRI response prediction.
In language modeling, \method{} improves next-token prediction by up to 25\%p over interpretable baselines, significantly narrowing the gap with black-box LLMs. In an fMRI setting, \method{} improves neural response prediction by 20\% and offers insight into the language selectivity of the brain.
\method{} represents a significant step toward uniting interpretability and performance across domains.
The code is available at \url{https://github.com/ejkim47/generalized-induction-head}.
\end{abstract}

\section{Introduction}

While modern transformer models have achieved impressive performance across a wide array of next-token prediction tasks~\citep{brown2020language,openai2023gpt4,dubey2024LLaMA}, these models remain black-boxes, limiting their use in real-world applications.
Their opacity is detrimental in fields such as neuroscience~\citep{jain2024computational} and social science~\citep{ziems2024can}, where trustworthy interpretation, specifically, token-level attribution that traces outputs back to input data, is often the end goal.
Their lack of transparency also hinders adoption in high-stakes applications such as medicine~\citep{thirunavukarasu2023large},
raising concerns around 
regulatory compliance, safety, and alignment~\citep{goodman2017european,amodei2016concrete,gabriel2020artificial,singh2024rethinking}.

As an alternative to these black-box models, 
interpretable models have been proposed for various tasks~\citep{rudin2022interpretable,mignan2019one,ha2021adaptive}, but they continue to struggle on the task of next-token prediction.
For example, in next-token prediction for natural language, the state-of-the-art interpretable model is \infinigram{}~\citep{liu2024infini}, which trails GPT-2 by 30\%p on the BabyLM dataset (see \cref{tab:peformance}).
Our analysis suggests that this performance gap stems from \infinigram{}’s inability to adapt to novel contexts or handle minor input variations such as typos and rephrasings.

We address this gap by proposing \methodlong{} (\method).
\method{} is inspired by the observation of ``induction heads'' in LLMs~\citep{olsson2022context,akyürek2024incontext} that support in-context learning by detecting and extending patterns in prior input.
In pre-trained LLMs, this behavior arises implicitly and is inferred through post-hoc approximations over dense internal states.
In contrast, \method{} is not a post-hoc tool for interpreting opaque systems, but an inherently interpretable system that explicitly models this behavior in a transparent and auditable manner.

\method{} is an interpretable retrieval-based framework that operates entirely within the model’s input context to retrieve suggestions for next-token completion.
It extends traditional exact matching by incorporating a lightweight fuzzy similarity function to match sequences that yield similar next-token distributions.
Importantly, this neural component is used solely for scoring similarity between input phrases rather than generating outputs.
The final next-token prediction is computed as a similarity-weighted distribution over tokens that follow the matched phrases, enabling each prediction to be directly attributed to specific input sequences, supporting full interpretability and human auditability. \method{} is designed as a standalone, model-agnostic module that supports both exact and fuzzy matching and can be integrated across modalities.

We first evaluate the performance and interpretability of \method{} in next-token prediction for language modeling.
We integrate \method{} into \infinigram{}, and \method{} improves next-token prediction accuracy by 25\%p over \infinigram{} using OpenWebText~\citep{Gokaslan2019OpenWeb} as a reference corpus, significantly narrowing the performance gap with GPT-2 (see \cref{tab:peformance}).

Second, we focus on a single, real-world neuroscience problem, deviating from a typical machine-learning conference paper.
Grounding in a neuroscience context allows us to avoid common pitfalls in evaluating interpretation methods~\citep{adebayo2018sanity,doshi2017roadmap} that seek to test ``interpretability'' in the abstract.
We find that when used to predict fMRI responses to language stimuli, \method{} yields a 20\% improvement over the state-of-the-art interpretable model (see \cref{tab:fmri_performance}), and its transparency enables the attribution of predicted neural responses in each region across the cortex to specific linguistic features.

Taken together, these results challenge the assumption that interpretability and predictive performance are fundamentally at odds, showing that reverse-engineered neural components can be leveraged to enhance transparency.
Importantly, our aim is not to claim parity with black-box LLMs, but to show that meaningful gains in predictive performance can be achieved without compromising transparency.

\section{Related Work}
\label{sec:related_work}
\paragraph{N-gram language models}
Early language modeling techniques revolved around n-gram models~\citep{martin2009speech,katz1987estimation}, which generally stored next-token probabilities in large tables learned from data~\citep{Brants2007LargeLM}.
While largely surpassed by neural LLMs, recent works have continued to improve n-gram LMs, e.g., by scaling up the n-gram reference data~\citep{ALLaMAnis2013MiningSC} and improving the n-gram probability representations using suffix arrays and suffix trees~\citep{Stehouwer2010UsingSA,Kennington2012SuffixTA,Shareghi2015CompactEA}.
This line of work culminated in \infinigram{}~\citep{liu2024infini}, which efficiently scales n-gram models to massive datasets and is the starting point for our work.

\paragraph{Bridging interpretable models and LLMs}
Some works have studied bridging n-gram models and LLMs.
For example, Khandelwal et al.~\citep{khandelwal2020generalization} interpolated neural LMs with an n-gram model and \mbox{Li et al.~\citep{li-etal-2022-residual}} trained a neural model to complement an n-gram model.
Other approaches augment black-box LMs with nonparametric components, such as $k$-nearest neighbors~\citep{khandelwal2020generalization,borgeaud2022improving}.
While these methods improve performance, they lack transparency in prediction behavior and token-level attribution.

Interpretable models have been proposed for simplified settings such as text classification. Some offer fully interpretable decision processes ~\citep{joulin2016bag,li2017neural,singh2023augmenting}, while others offer partial interpretability by approximating model behavior with natural language concepts~\citep{yang2023language,sun2024concept,singh2023tree,feng2024bayesian,mcinerney2023chill}. However, these approaches are not designed for open-ended generation.

In parallel, there has been a recent surge of interest in mechanistic interpretability, which seeks to understand what mechanisms are learned by transformer-based LLMs~\citep{rai2024practical,meng2022locating,elhage2021mathematical,saphra2024mechanistic}.
This line of work identified induction heads in toy LLM models~\citep{olsson2022context} as well as large-scale pre-trained LLMs~\citep{wang2022interpretability,akyürek2024incontext}.
Despite these efforts, frameworks to make these findings useful in real settings remain underexplored.

\begin{figure}[t]
\centering
\includegraphics[width=0.75\linewidth]{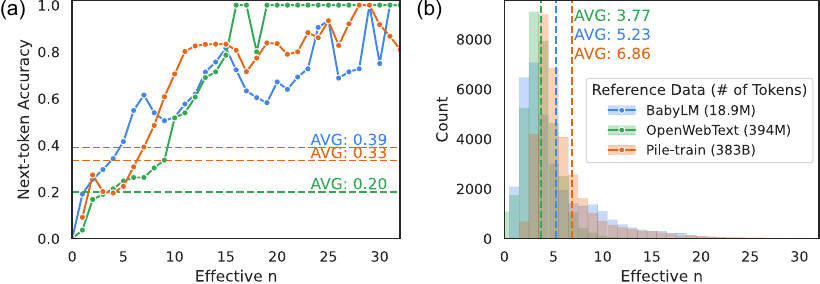} 
  \caption{
  Performance on the BabyLM dataset with \infinigram{} built from various reference datasets.
  (a) Next-token prediction accuracy by effective $n$, with the dashed line indicating the average.
  (b) The histogram of the count for each effective $n$.
  }
  \label{fig:result_need_incontext}
\end{figure}

\paragraph{Natural language representations in fMRI}
In recent years, predicting brain responses  to natural language using LLM representations has become common in the field of language neuroscience~\citep{jain2018incorporating,wehbe_aligning_2014,schrimpf2021neural, TONEVANEURIPS2019, goldstein_shared_2022}.
Predictive ``encoding models'' have been used to explore the relative contributions of syntax, semantics, and discourse to neural activity \citep{wu_complete_2006,caucheteux_disentangling_2021,
kauf2024lexical, reddy2021can,
pasquiou2023information, aw2022training, 
kumar_reconstructing_2022, oota2023joint} and to study the cortical organization of language timescales \citep{JAINNEURIPS2020, chen2024cortical}, sometimes making use of LLMs to help annotate and generate stimuli~\cite{tuckute2023driving,benara2024crafting,singh2025evaluating,antonello2024generative}. 
However, these models largely operate as black boxes. While they reveal which language representations best predict neural activity, they offer limited insight into where in the cortex these features exert their influence or when in the linguistic stimulus they become relevant.

Separately, behavioral studies have examined how humans recall and process repeated text ~\citep{baddeley1992working,tzeng1973positive,amlund1986repetitive,miles2006verbatim} and how similar recall patterns emerge in LLMs~\citep{vaidya2023humans,pink2024assessing}. Yet the cortical mechanisms involved in contextual recall remain unclear, motivating our investigation through interpretable modeling.

\section{Methods}

We begin by discussing \infinigram{},
the scalable n-gram method for interpretable next-token prediction (\cref{subsec:infinigram}), and introduce \methodlong{} (\method) (\cref{subsec:in_context_method}).
In describing the method, we focus on the familiar scenario of next-token prediction for language modeling, but note that the method straightforwardly generalizes to generic next-token prediction tasks (e.g., fMRI responses, time series, video frames).
We later show how \method{} improves interpretable next-token prediction by integrating it with \infinigram{} for language modeling (\cref{sec:lang_model}) and combining it with linear regression for fMRI response prediction (\cref{subsec:fmri}).

\subsection{Preliminaries and Motivation: \infinigram{}}
\label{subsec:infinigram}
Given an input text sequence, \infinigram{}~\citep{liu2024infini} searches a reference corpus for the longest exact suffix match to the input, then calculates the next-token distribution based on the token following each of the matches.
This search is made efficient by building large-scale suffix arrays that can scale to trillions of reference tokens.
The length of the longest match is referred to as the \textit{effective} $n$, with the accuracy of the estimated probabilities increasing as the \textit{effective} $n$ becomes larger.

\infinigram{} is limited by its reliance on exact matches, which becomes problematic under distribution shifts between the input and reference corpus. For instance, when evaluating on the BabyLM\footnote{\scriptsize \url{https://babylm.github.io/}}~\citep{warstadt2023findings} test set, \infinigram{} built on larger corpora, such as OpenWebText~\citep{Gokaslan2019OpenWeb}, shows lower performance and, on average, has fewer instances of higher effective $n$ compared to the model built on the BabyLM (\cref{fig:result_need_incontext}). With far larger corpora like Pile-train~\citep{gao2020pile}, \infinigram{} is able to increase the number of instances with a high effective $n$, resulting in improved performance. However, \infinigram{} built on BabyLM, which contains only 0.005\% of the tokens found in Pile-train, still achieves the highest performance. This highlights the difficulty \infinigram{} faces when there is a substantial gap between the reference corpus and the input prompt, making it hard to find matching cases with a large effective $n$. We address this limitation with the concept of the induction head.

\subsection{Building a Generalized Induction Head}
\label{subsec:in_context_method}

LLMs excel at in-context learning by capturing the statistical distribution of tokens in a given context. 
One key mechanism enabling this capability is the induction head, a critical component in LLMs responsible for recognizing and extending repeated sequences~\citep{olsson2022context,akyürek2024incontext,wang2022interpretability}. Induction heads operate by detecting prior occurrences of a token sequence and leveraging this recurrence for next-token prediction (e.g., given [A][B] ... [A], the model predicts [B]). However, these heads emerge implicitly within LLMs, making their operation difficult to interpret and control.

To this end, we introduce \methodlong{} (\method{}) that explicitly models this behavior in a structured, interpretable manner.
It functions like \infinigram{} but is restricted to the input context. \method{} treats the end of the context as a query, searches for the best match within the context and takes the token following the match as the next-token prediction.

\paragraph{What constitutes a ``good match''?}
When identifying n-gram-level matches in context, exact matching can perform well if a high effective $n$ is guaranteed (\Cref{subsec:infinigram}),
but it can be overly restrictive to minor changes such as rephrasings or typos.
To remedy this, we allow for fuzzy n-gram matching, which makes the model more robust to minor changes.
Since the fuzzy matching is performed at the level of n-grams, predictions remain interpretable and auditable by a human.

Fuzzy matching requires appropriately computing the similarity between sequences.
While similarity can be defined in many ways, in building an induction head, we desire two sequences to be similar if they yield similar next-token distributions. To quantify this, we define the similarity between two sequences, $x_1$ and $x_2$, for fuzzy matching using Jensen–Shannon divergence (JSD):
\begin{equation}\label{eq:similarity}
s(x_1, x_2)=\exp\left(-\text{JSD}\left(P_{\text{next}}(x_1), P_{\text{next}}(x_2)\right)\right),
\end{equation}
where $P_{\text{next}}(\cdot)$ is the estimated next-token probability distribution for a given sequence.

\begin{figure*}[t]
    \centering
    \includegraphics[width=\textwidth]{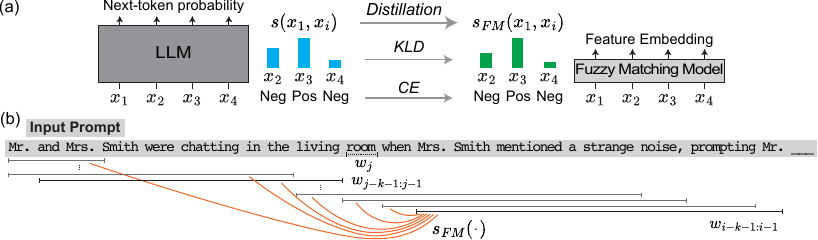}
    \caption{(a) Overview of training \fuzzyslmlong{} via distillation from a pretrained LLM. (b) Calculation of sequence similarity within the input prompt for next-token prediction.
    }
    \label{fig:method}
\end{figure*}

\paragraph{Computing $s$ efficiently}
One approach for computing $s$ would be to use a pre-trained LLM to obtain $P_{\text{next}}$, but this can be computationally expensive.
Instead, we develop a small \fuzzyslmlong{}, which consists of a few transformer layers and is trained via knowledge distillation from existing LLMs.
This model is designed to output feature embeddings that facilitate the calculation of next token probabilities for similarity assessments. With the \fuzzyslmlong{}, the similarity between $x_1$ and $x_2$, whose feature embeddings from the model are $e_1$ and $e_2$, is obtained as follows:
\begin{equation}\label{eq:similarity_fuzzy}
s_{\text{FM}}(x_1, x_2)=\exp\left(-\left(1 - \text{CosSim}\left(e_1, e_2\right)\right)/T\right),
\end{equation}
where $T$ is a temperature, which is set to 0.1.
The \fuzzyslmlong{} is trained with a combination of Cross Entropy (CE) loss and reverse Kullback-Leibler divergence (KLD) loss (\cref{fig:method}(a)). In each training batch, we generate similarity pairs from randomly sampled sequences. 
The CE loss aids in identifying the most similar pairs. 
The reverse KLD loss guides the model to follow the overall similarity distribution, ensuring that close pairs receive high scores while distant pairs receive low scores.
Further details can be found in Appendix~\ref{sec:fuzzyslm_appendix}.

\paragraph{Predicting the next token}
Given the similarity scoring function $s_{\text{FM}}$,
we construct an induction head that yields the predicted next-token probability distribution $P_{\text{induction}}^{\text{(fuzzy)}}$ given an input sequence $x$.
To achieve this, we identify matches for the end of $x$, $w_{:i-1}$, using a sliding window of size $k$ (\cref{fig:method}(b)).
We then count the occurrence of each token $w_i$ in the vocabulary set $\mathcal{V}$ following these matches and normalize to obtain the next-token probability:
\begin{equation}\label{eq:cnt}
P_{\text{induction}}^{\text{(fuzzy)}}(w_{:i-1}w_i|x)=\frac{c_\text{fuzzy}(w_{i-k-1:i-1}w_i|x)}{\sum_{w_j\in\mathcal{V}}{c_\text{fuzzy}(w_{i-k-1:i-1}w_j|x)}},
\end{equation}
\begin{equation}
\text{where } c_\text{fuzzy}(w_{i-k-1:i-1}w_i|x)=\sum_{w_{j-k-1:j}\subset x}{\mathds{1}_{w_j=w_i}s_{\text{FM}}\left(w_{j-k-1:j-1}, w_{i-k-1:i-1}\right)}.
\end{equation}
This similarity score serves as a floating count for the next token.
In cases where the sequences $x_1$ and $x_2$ are exactly matched, as in the case of \infinigram{}, we have $s_{\text{FM}}(x_1, x_2)=1$, which is equivalent to increasing the count by one. The window size $k$ specifies the number of tokens to be considered in fuzzy matching.

\subsection{Prediction of Generalized Induction head Model}
\begin{figure*}[!t]
    \centering
    \includegraphics[width=\textwidth]{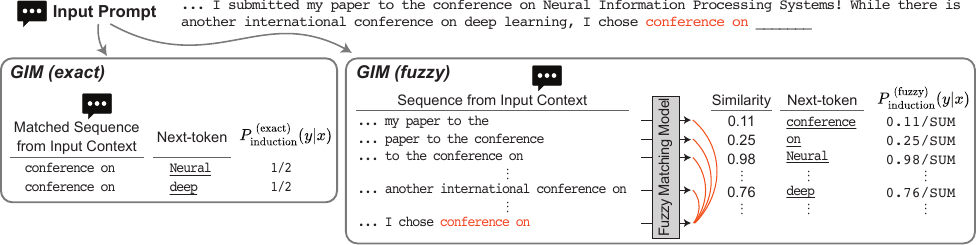}
    \caption{Overview of the \method{} pipeline.
    \method{} predicts the next token by efficiently searching for potential next-token completions in the input context with either exact or fuzzy matching.
    }
    \label{fig:overview}
\end{figure*}

By employing both the \infinigram{} algorithm and \fuzzyslmlong{}, \method{} searches for the most relevant match—either exact or fuzzy—within the preceding tokens given a query at the end of the context (\cref{fig:overview}). Once a match is identified, it retrieves the token that followed the prior occurrence as the next-token prediction. By explicitly modeling this process, our method provides a transparent and controllable alternative to implicit in-context learning mechanisms in LLMs.

\section{Results: Next-token Prediction for Language Modeling}\label{sec:lang_model}

\subsection{Experimental Setup}\label{subsec:lang_model_setup}

\paragraph{Datasets \& evaluation} We use 4 text datasets for evaluation: BabyLM~\citep{warstadt2023findings}, OpenWebText~\citep{Gokaslan2019OpenWeb}, Pile~\citep{gao2020pile}, and FineWeb~(\citep{penedo2024fineweb}; \texttt{sample-10BT} subset), using some as the reference corpus and some as test datasets (\cref{tab:peformance}).
When testing, we report performance on 100k sequences randomly sampled with a context length of 1024 and a stride of 512~\citep{liu2024infini,khandelwal2020generalization}.\footnote{The BabyLM test set contains fewer than 100k sequences, yielding approximately 32k and 34k cases for the GPT-2 and LLaMA-2 tokenizers, respectively.}
We evaluate next-token prediction via accuracy, i.e. whether the top-predicted token was the correct token.\footnote{We do not use perplexity, as the sparse next-token predictions from n-gram models often assign zero probability to the top-ranked token, resulting in undefined or extremely high perplexity scores~\citep{liu2024infini}.}

\paragraph{Baselines}
We compare against \infinigram{} as our sole baseline, as it is the state-of-the-art n-gram model and the only fully interpretable model with token-level attribution for generation.
We found that it consistently outperformed prior interpretable language models, e.g., a standard 5-gram model based on OpenWebText achieves 26.4\% accuracy in next-token prediction on the Pile-val, lower than \infinigram{}'s 27.1\%. For a detailed discussion of interpretable frameworks, please refer to \cref{sec:related_work}.

\paragraph{Integrating \method{} with \infinigram{}}

For language modeling, we integrate \method{} with \infinigram{}, enabling the use of both reference corpus statistics and in-context distributions:
\begin{equation}
\label{eq:main}
P(y|x) = 
\begin{cases}
\vspace{5pt}
P_{\infty}^{\text{ (exact)}}(y|x) & n_{\infty} > n_x  ~\text{and}~ n_{\infty} >\tau, \\
\vspace{5pt}
P_{\text{induction}}^{\text{ (exact)}}(y|x) & n_x \geq n_{\infty} ~\text{and}~ n_x >\tau,\\ 
P_{\text{induction}}^{\text{ (fuzzy)}}(y|x) & \text{Otherwise,}
\end{cases}
\end{equation}
where $n_\infty$ and $n_x$ are the effective $n$ when matching from a reference corpus or the input context, respectively.
When these values are low, fuzzy matching is employed to compensate for the limited effective $n$.
When the effective $n$ values from both the input context and reference corpus are equal, the estimate from the input context is prioritized.
The hyperparameter $\tau$ determines how frequently exact matching is used over fuzzy matching; we set $\tau$ to 8 and 9 for the GPT-2 and LLaMA-2 tokenizers, respectively, based on cross-validation results (see \cref{sec:effective_n_thres} for details).

\begin{table*}[t]
\caption{Next-token prediction accuracy (\%) for language modeling. Gray shading represents alignment between the reference corpus and the test dataset.}
\label{tab:peformance}
\small
\begin{center}
\begin{adjustbox}{max width = 0.97\textwidth}
\begin{tabular}{lllllll}\toprule
\multicolumn{2}{c}{Reference Corpus} &\multirow{2}{*}{Model} &\multicolumn{3}{c}{Test Dataset} \\\cmidrule{1-2}\cmidrule{4-6}
\;\;\;Type &\# of Tokens & &BabyLM-test &FineWeb &Pile-val \\\midrule
\hspace{-4pt}\multirow{2}{*}[1ex]{\textbf{Tokenizer: GPT-2}} \\
\;\;\;- &- &\exactinduction{} &36.7 &17.2 &37.0 \\
\;\;\;- &- &\fuzzyinduction{} &41.1 &25.2 &38.7 \\
\;\;\;\multirow{2}{*}{BabyLM-dev} &\multirow{2}{*}{17.4M} &\infinigram{} & \cellcolor{gray!20}37.6 &14.7 &16.0 \\
\;\;\;& &~+\textbf{\method{}} &\cellcolor{gray!20}42.2 {\tiny(+4.6)} &25.3 {\tiny(+10.6)} &40.0 {\tiny(+24.0)} \\
\;\;\;\multirow{2}{*}{Pile-val} &\multirow{2}{*}{383M} &\infinigram{} &16.6 &20.1 &- \\
\;\;\;& &~+\textbf{\method{}} &41.5 {\tiny(+24.9)} &25.5 {\tiny(+5.4)} &- \\
\;\;\;\multirow{2}{*}{OpenWebText} &\multirow{2}{*}{9.04B} &\infinigram{} &16.7 &25.5 &22.7 \\
\;\;\;& &~+\textbf{\method{}} &41.8 {\tiny(+25.1)} &27.2 {\tiny(+1.7)} &42.7 {\tiny(+20.0)} \\
\;\;\;Unknown & $\sim$10B &LLM (GPT-2) &46.9 &39.0 &52.3 \\
\midrule
\hspace{-4pt}\multirow{2}{*}[1ex]{\textbf{Tokenizer: LLaMA-2}} \\
\;\;\;- &- &\exactinduction{} &37.0 &19.6 &32.6 \\
\;\;\;- &- &\fuzzyinduction{} &42.7 &28.3 &38.5 \\
\;\;\;\multirow{2}{*}{BabyLM-dev} &\multirow{2}{*}{18.9M} &\infinigram{} &\cellcolor{gray!20}39.0 &17.1 &13.2 \\
& &~+\textbf{\method{}} &\cellcolor{gray!20}43.1 {\tiny(+4.1)} &28.6 {\tiny(+11.5)} &39.6 {\tiny(+26.4)} \\
\;\;\;\multirow{2}{*}{Pile-val} &\multirow{2}{*}{394M} &\infinigram{} &19.0 &24.1 &- \\
& &~+\textbf{\method{}} &42.9 {\tiny(+23.9)} &28.4 {\tiny(+4.3)} &- \\
\;\;\;\multirow{2}{*}{OpenWebText} &\multirow{2}{*}{10.3B} &\infinigram{} &20.1 &29.5 &27.1 \\
& &~+\textbf{\method{}} &43.2 {\tiny(+23.1)} &30.3 {\tiny(+0.8)} &42.1 {\tiny(+15.0)} \\
\;\;\;\multirow{2}{*}{Pile-train} &\multirow{2}{*}{383B} &\infinigram{} &33.5 &39.3 &\cellcolor{gray!20}49.2 \\
& &~+\textbf{\method{}} &49.4 {\tiny(+15.9)} &38.0 {\tiny(-1.3)} &\cellcolor{gray!20}50.3 {\tiny(+1.1)} \\
\;\;\;Unknown & $\sim$2T &LLM (LLaMA2-7B) &62.2 &57.1 &64.4 \\
\bottomrule
\end{tabular}
\end{adjustbox}
\end{center}
\end{table*}

\subsection{Improving Next-token Prediction Accuracy with Contextualization}\label{subsec:next-token_prediction_results}

\begin{figure}[t]
    \centering
    \includegraphics[width=0.8\linewidth]{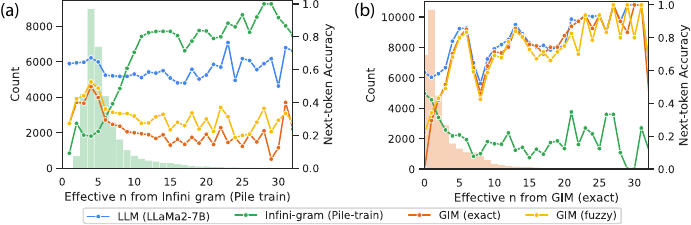}
    \caption{Comparison of next token prediction accuracy on BabyLM-test dataset, depending on effective $n$ from (b) \infinigram{} and (b) \exactinduction{}.
    LLaMA-2 tokenizer is used.}
    \label{fig:result_effectiven_plot}
    \vspace{3pt}
\end{figure}

\paragraph{Prediction performance of in-context matching}
\method{} relies solely on the input context to predict the next token (limited to 1024 tokens in our evaluation). \Cref{tab:peformance} shows that, despite this, \exactinduction{} can outperform \infinigram{}—which uses the OpenWebText dataset as a reference corpus, comprising approximately 10B tokens when tokenized with LLaMA-2 and 9.04B with GPT-2—by a margin of 5.5\%p to 20\%p on the BabyLM and Pile datasets. \infinigram{} using BabyLM-dev as the reference corpus slightly outperforms \exactinduction{} on the BabyLM-test, with performance gaps of 0.9\%p and 2.0\%p for the GPT-2 and LLaMA-2 tokenizers, respectively, under the aligned setting of reference corpus and input context.
As shown in \cref{fig:result_effectiven_plot}(a), \infinigram{} (green) performs better in cases with a high effective $n$, even surpassing LLM (blue).
However, significantly more cases have a low effective $n$ (histogram), where \exactinduction{} (orange) outperforms \infinigram{}.
This finding underscores that in-context matching reflects the input query's distribution, leading to more accurate next-token predictions than reference matching, even when the reference corpus contains abundant tokens, especially under distribution shifts between the reference corpus and the test input.

\paragraph{Prediction improvements from \method{}}
\fuzzyinduction{}, using \fuzzyslmlong{}, consistently outperforms \exactinduction{} with a margin of 1.7\%p to 8.7\%p (\cref{tab:peformance}). This improvement is particularly evident in cases with low effective $n$. As illustrated in \cref{fig:result_effectiven_plot}(b), the majority of cases within the input context have low effective $n$ (histogram), indicating that finding exactly matched long sequences within the limited amount of tokens is challenging. Fuzzy matching helps to provide better estimations for next-token predictions in these scenarios. Specifically, when the effective $n$ is less than 3, \fuzzyinduction{} (yellow) demonstrates better performance than \exactinduction{} (orange). Since many cases fall into this range, the overall accuracy of \fuzzyinduction{} is higher.

The improvements achieved through the use of induction and fuzzy matching enable \infinigram{} with \method{} to outperform \infinigram{} built on 383B tokens, improving performance by up to 15.9\%p. While enlarging the reference corpus boosts performance, \method{} offers a more efficient alternative to scaling from 10.3B to 383B tokens—a 38-fold increase. Moreover, \method{} is a complementary approach that can be applied orthogonally to \infinigram{}, regardless of the size of the reference corpus.

\begin{figure}[t]
    \centering
    \includegraphics[width=\textwidth]{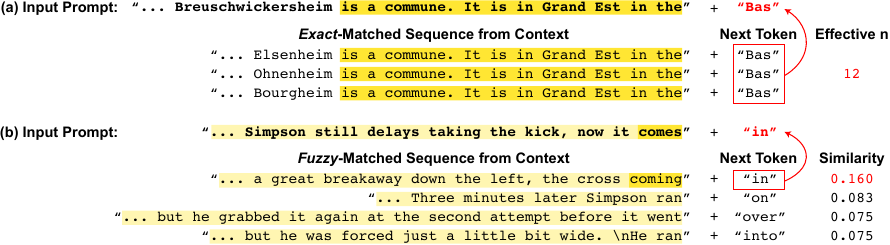}
    \caption{\method{}'s token-level attribution by tracing predictions to (a) exact or (b) fuzzy matches in prior context. Yellow highlight shows the match, and red box marks the source of the final prediction.}
    \label{fig:explanation_main}
    \vspace{2pt} 
\end{figure}

\subsection{Qualitative Example of GIM Prediction}
\cref{fig:explanation_main} shows examples of explanations provided by \method{}. In the first case, the prompt exactly matches a 12-gram in the context, so \method{} follows it to predict the next token. In the second, no exact match exists for the prompt ending in ``comes'', but \method{} finds the most similar sequence ending in ``coming'' and follows it for prediction. These cases illustrate how \method{} predicts from retrieved sequences, with transparency into which tokens contribute and how they are combined.

\section{Results: Next-token prediction for fMRI Responses to Natural Language}\label{subsec:fmri}

Understanding how and where semantic information is represented across the human brain is a central objective in neuroscience. In this work, we extend prior modeling frameworks that learn mappings between natural language stimuli and corresponding neural responses across voxels, which are small three-dimensional regions of the brain.~\citep{huth2016natural,jain2018incorporating}

\FloatBarrier
\subsection{Experimental setup}\label{subsec:fmri_experimental_setup}
\begin{wrapfigure}{r}{0.51\textwidth}
    \vspace{-12pt}   
    \centering
    \includegraphics[width=\linewidth]{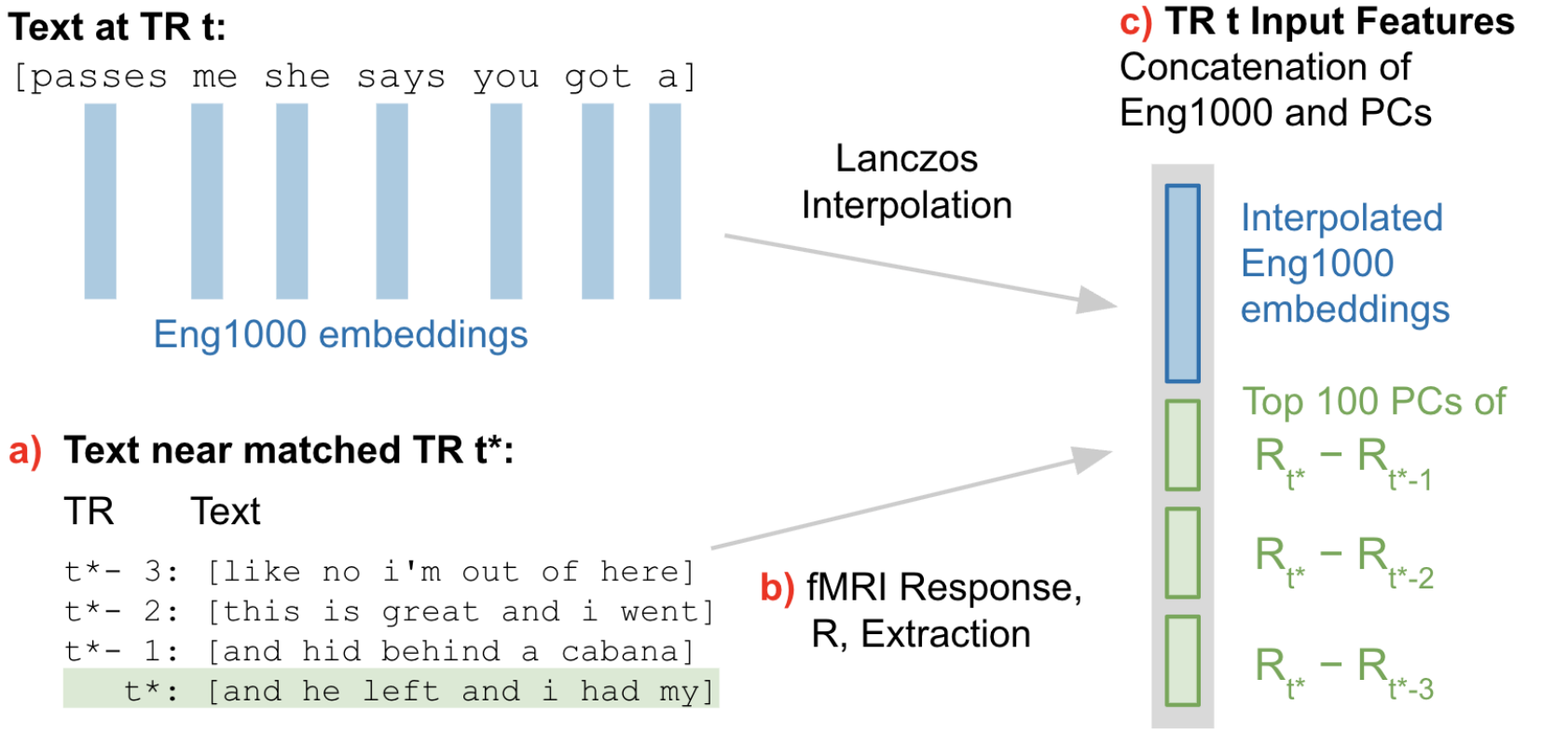}
    \caption{
    fMRI feature construction. 
    (a) At each TR $t$, we retrieve a prior TR $t^*$ with various matching methods.
    (b) We extract the top 100 principal components (PCs) of neural response changes before $t^*$. 
    (c) These are concatenated to interpolated Eng1000 embeddings for fMRI signal prediction at TR $t$.}
    \label{fig:fmri_method}
    \vspace{-10pt}
\end{wrapfigure}
We analyzed publicly available data\footnote{\scriptsize \url{https://github.com/OpenNeuroDatasets/ds003020}} from \citep{lebel2022natural} and \citep{tang2023semantic}, in which three human participants listened to 20+ hours of English-laguage podcast narratives while their fMRI responses were recorded across 95,556 cortical voxels.
Our goal was to predict the brain response of each voxel from the language input heard by the participant\footnote{We report results for subject UTS03 due to high fMRI data quality, including superior repeatability, minimal motion, and strong encoding model performance~\citep{lebel2022natural}.}. We extracted text embeddings from the input story, then fit linear models to map these embeddings to fMRI responses on the training split (24 stories), and evaluated performance on the test split (2 stories) using bootstrapped ridge regression.
Embeddings are extracted in various ways (described below) for each word in the input, and then interpolated to make predictions for the fMRI data that is recorded at 2-second time of repetition (TR) intervals.
To model temporal delays in the fMRI signal, we add 4 time-lagged duplicates of the input features.
See more fMRI details in \cref{subsec:fmri_appendix}.

\paragraph{fMRI prediction baselines}
We use Eng1000 as our primary baseline, the state-of-the-art interpretable model for predicting fMRI responses to narrative stories from a seminal study of language selectivity~\citep{huth2016natural}.
Each element in an Eng1000 embedding corresponds to a co-occurrence statistic with a different word.
We also compare against LLaMA2-70B~\citep{touvron2023llama2} embeddings, which achieve the highest performance on this task~\citep{antonello2024scaling} but are not interpretable.
LLaMA embeddings are extracted with a 16-word sliding window, using the final-layer embedding of the last token in each window.

\paragraph{\method{} for fMRI prediction}
We construct our \method{} for fMRI by searching the preceding story text in an fMRI session for semantic matches and retrieving the changes in the recorded brain response that follows each match. 
Specifically, to predict the fMRI response, $R_{t}$, for the TR $t$, we first find the TR $t^{*}$ for which the text input yields the highest cosine similarity to the next-token distribution of the text input at TR $t-1$.
Next, we isolate the change in fMRI responses following TR $t^{*}$: we take the difference in the top 100 principal components of the response
$R_{t^{*}}-R_{t^{*} - 1}$ and use them as features.
To deal with potential time delays in the fMRI signal, we additionally concatenate these features with the top 100 principal components of $R_{t^{*}} - R_{t^{*} - 2}$ and $R_{t^{*}} - R_{t^{*} - 3}$.
These features, along with the interpolated Eng1000 embeddings, form the full input to the linear model predicting fMRI response at TR $t$ (see \cref{fig:fmri_method}).
When constructing the \method{} for fMRI, we search over the most recent 1024 words and their corresponding fMRI responses.
To measure similarity between two texts, we use the predicted next-word distributions yielded by \exactinduction{} in the input context ($P_\text{induction}^{ \text{ (exact)}}$ in \cref{eq:main}), which we call \textit{\method{} matching}.

\paragraph{Baseline matching methods} We compare GIM matching against three baseline matching strategies. 
First, we use the predicted next-word distributions yielded by exact n-gram matching in the 10B-token OpenWebText reference corpus ($P_{\infty}^{ \text{ (exact)}}$ in \cref{eq:main}), which we call \textit{\infinigram{} matching}.
Second, \textit{Random matching} selects a random preceding TR as a match.
Third, \textit{Naive n-gram matching} searches for an exact match to the most recent 4-word n-gram in the input context, without relying on predicted next-word distributions that our \method{} matching method relies on.
\cref{tab:fuzzy_matching} shows additional experiments with fuzzy matching methods that show little performance gain, likely due to noise and temporal smoothing in fMRI signals that diminishes the advantage of fuzzy matching.

\subsection{Prediction improvements from \method{} matching}

\begin{figure}[t]
\centering
\begin{minipage}[c]{0.47\textwidth}
\setlength{\tabcolsep}{2.5pt}
\centering
\captionof{table}{fMRI test prediction performance for different models.
Black-box encodings use LLaMA-2.
Error bars show 95\% CI.}
\small
\label{tab:fmri_performance}
\begin{center}
\begin{adjustbox}{max width = \textwidth}
\begin{tabular}{lcc}\toprule
\multirow{3}{*}{Feature Model} & \multicolumn{2}{c}{Mean Correlation} \\\cmidrule{2-3}
& All & Top 10\% \\
& Voxels & Voxels \\\midrule
Eng1000 & 0.072\err{0.0004} & 0.220\err{0.0012} \\
\midrule
~+ Random matching & 0.069\err{0.0003} & 0.197\err{0.0012} \\
~+ Naive ngram matching  & 0.068\err{0.0003} & 0.194\err{0.0012} \\
~+ \infinigram{} matching  & 0.069\err{0.0003} & 0.200\err{0.0012} \\
~+ \textbf{\method{} matching} & \textbf{0.087\err{0.0005}} & \textbf{0.265\err{0.0011}}\\\midrule
Black-box encodings & 0.096\err{0.0005} & 0.268\err{0.0013} \\\bottomrule
\end{tabular}
\end{adjustbox}
\end{center}
\end{minipage}%
\hfill%
\begin{minipage}[c]{0.49\textwidth}
\centering
\centering
\includegraphics[width=\linewidth]{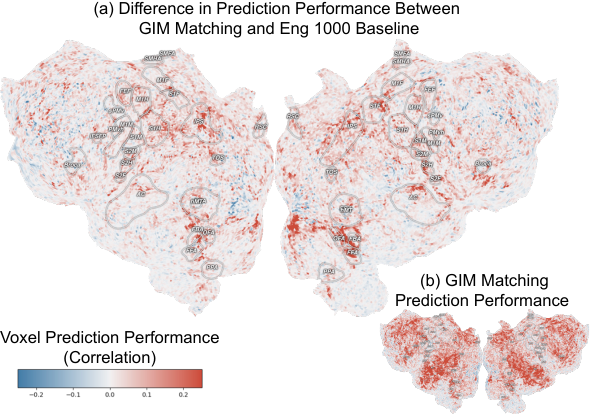}
\caption{(a) Difference in the correlation performance between the GIM matching and the Eng1000 baseline, visualized across the cortex. (b) Correlation performance of GIM matching.}
\label{fig:flatmap_diff}
\end{minipage}
\end{figure}

Table~\ref{tab:fmri_performance} shows the average correlation values across all voxels for each similarity model.
Eng1000, the primary interpretable baseline, achieved a mean test correlation of 0.072.
In contrast, \method{} matching achieves a mean correlation of 0.087, a 20\% improvement over Eng1000.
When predicting the top-10\% of voxels, \method{} Matching achieves a mean correlation of 0.265, again a 20\% improvement over Eng1000, and only 1\% lower than the black-box LLaMA-2 model (mean correlation 0.268).
In contrast, other matching-based baselines are unable to improve over Eng1000:
The Naive n-gram matching baseline achieves a correlation of 0.068, and random matching achieves a correlation of 0.069.

\cref{fig:flatmap_diff} visualizes voxel-wise differences in test correlation performance between \method{} matching and the Eng1000 baseline across the cortex. In line with prior studies linking model performance to functional localization~\citep{goldstein_shared_2022, schrimpf2021neural, huth2016natural, jain2018incorporating}, \method{} significantly improves prediction in regions highted in red such as the Occipital Face Area (OFA) and Intraparietal Sulcus (IPS). These gains reflect \method{}’s use of contextual input, in contrast to the static embeddings used by Eng1000, and suggest that these highlighted regions may contribute to contextual language processing.

\paragraph{Describing improvements from \method}

\begin{figure}[t]
    \centering
    \includegraphics[width=\textwidth]{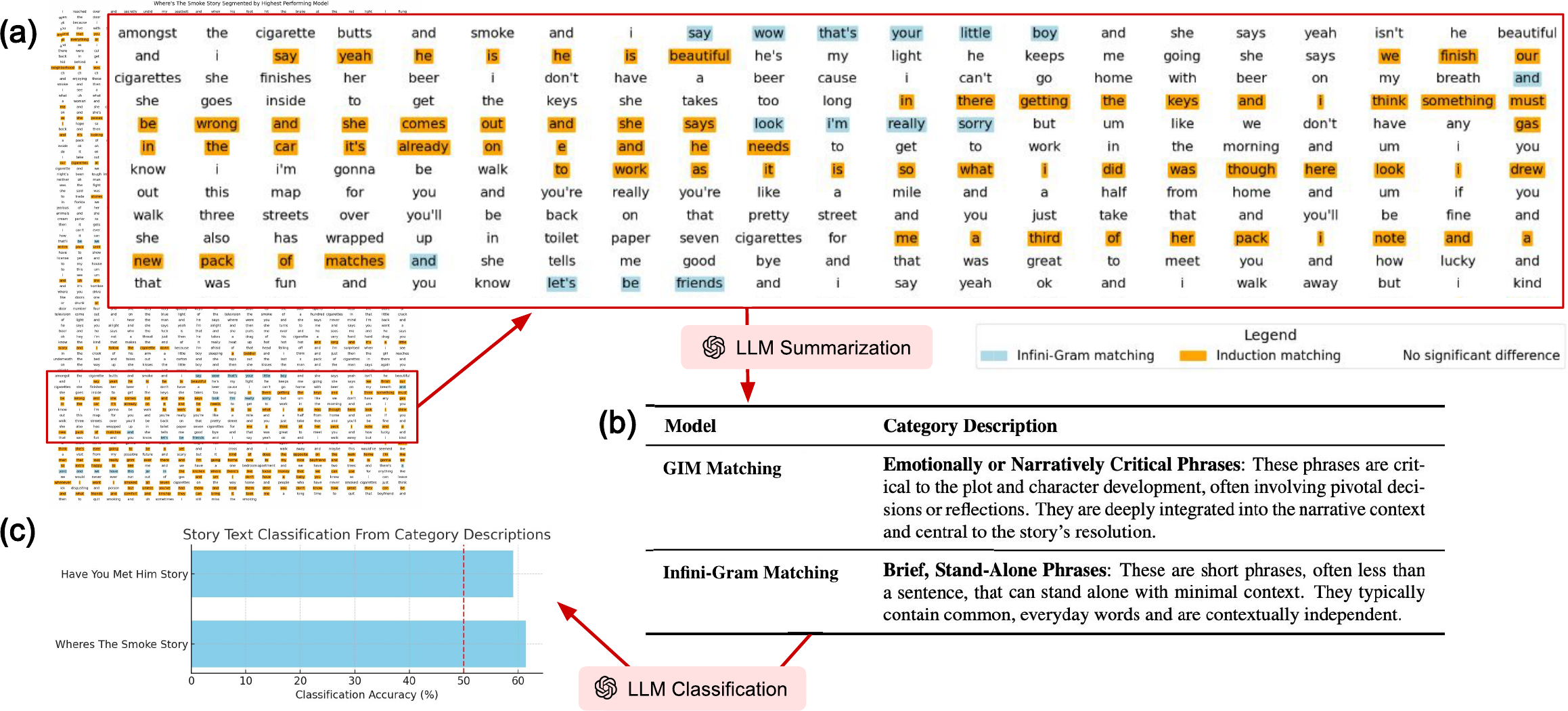}
    \caption{Qualitative illustration of how \method{} and \infinigram{} matching improve performance.
    (a) Highlighted phrases in the input story where the model outperforms the baseline.
    (b) Summary of the highlighted phrases by an LLM to characterize each matching method.
    (c) Classification of highlighted phrases in test stories based on the LLM-generated summaries.
    }
    \label{figs:category_descriptions}
\end{figure}

To understand the improvements provided by matching, we summarize the text for inputs where each matching procedure (\method{} and \infinigram{}) performs well.
We use an LLM to do the summarization, following recent works in LLM interpretability~\citep{zhong2022describing,dunlap2024describing}.
We first identify phrases in the input story where a model's performance (average absolute error across voxels) exceeds the baseline performance by more than one standard deviation (see an example in \cref{figs:category_descriptions}).
Then, we prompt GPT-4 (\citep{openai2023gpt4}; \texttt{gpt-4-0613}) to generate descriptions for these phrases. 

\cref{figs:category_descriptions} presents the unedited LLM descriptions\footnote{Irrelevant preceding text such as ``Sure here is the answer'' is removed from the response.}.
\method{} matching is described as capturing \textit{Emotionally or Narratively Critical Phrases}, aligning with the idea that induction improves performance by tracking local context in a story, e.g., phrases that ``are critical to the plot and character development''.
In contrast, \infinigram{} matching is described as capturing \textit{Brief, Stand-Alone Phrases}, matching the intuition that \infinigram{} excels in capturing context that is not story specific, but ``can stand alone with minimal context''.
To test these descriptions, we prompt GPT-4 to classify the identified phrases in two test stories using only the descriptions.
This yields 61\% accuracy, a moderate but significant improvement over chance (binomial test $p = 0.032$). See all phrases and prompts in \cref{subsec:fmri_appendix}.

\section{Discussion}

\method{} constitutes a significant step toward building mechanistically interpretable language models inspired by pre-trained LLMs. Unlike black-box models or partially interpretable approaches, \method{} provides full transparency in next-token prediction while substantially narrowing the performance gap between interpretable and black-box architectures across two diverse domains. Importantly, \method{} is not a general-purpose LLM or a tool to decode its internals; it isolates and reimplements a single observed capability, induction via repetition, as a fully interpretable module. This shows that high-performance behaviors implicitly learned by LLMs can be transparently reconstructed to advance performance in the interpretable modeling space.
The transparency of \method{} makes it well-suited for language modeling scenarios that require complete auditing, such as analyzing scientific texts or medical notes~\citep{yang2023large}. \method{}'s transparency also supports neuroscience research, as the fMRI analyses conducted here are a suggestive starting point for understanding how context is stored and recalled in the human cortex. GIM can further serve as a testbed for analyzing how context modulates the recall of specific semantic categories, like people and places, across the cortex, extending prior work with static embeddings~\citep{huth2016natural}.
Additionally, improvements from \method{} Matching may help build encoding models that can more rapidly adapt to local context, which can be used in downstream applications such as brain decoding~\citep{tang2023semantic} or brain-computer interfaces~\citep{nicolas2012brain}.

\method{} shows limited gains when the input context is short or uninformative. Its modular design enables the available context to be expanded through retrieval-augmented generation~\citep{wu2024retrieval} or external memory. Like kNN-LMs~\citep{geng2024great}, \method{}’s n-gram-based reasoning also struggles with tasks requiring deeper reasoning. Future work may explore hybrid approaches that pair \method{} with black-box models for better trade-offs. Our speculative decoding setup, where \method{} serves as a transparent draft generator verified by a larger LLM (\cref{sec:sp}), illustrates one example in this direction.
Another promising direction is expanding \method{} beyond induction heads, integrating additional mechanistic components such as indirect object identifiers~\citep{wang2022interpretability}, numerical representations~\citep{engels2024not}, retrieval heads~\citep{wu2024retrieval}, 
iteration heads~\citep{cabannes2024iteration},
concept-level induction heads~\citep{feucht2025dual},
instruction-following heads~\citep{zhang2023tell}, or interpretable LLM submodules~\citep{bills2023language,singh2023explainingmodules,bricken2023monosemanticity}.
Finally, \method{}'s ability to model context-dependent patterns makes it well-suited for other sequential domains that require interpretability, e.g.,
it could be extended to study long-range dependencies in electronic health records~\citep{renc2024zero},
audio/speech models~\citep{caucheteux2023evidence,shimizu2025interpretable},
genomics~\citep{avsec2021effective} or financial time-series analysis~\citep{lim2021temporal}.

\section*{Acknowledgements}
We would like to thank Lucas Liu, Ziyang Wu, and Paul Smolensky for insightful discussions and feedback throughout this work, which significantly contributed to the development of our ideas. This work was supported by Institute of Information \& communications Technology Planning \& Evaluation (IITP) grant funded by the Korea government (MSIT) [No.RS-2021-II211343, Artificial Intelligence Graduate School Program (Seoul National University), RS-2022-II220959, No.RS-2025-02263754, Human-Centric Embodied AI Agents with Autonomous Decision-Making], the BK21 FOUR program of the Education and Research Program for Future ICT Pioneers, Seoul National University in 2025, and the National Research Foundation of Korea (NRF) grant funded by the Korea government (MSIT) (No. 2022R1A3B1077720, 2022R1A5A708390811).

{
\small
    \bibliography{refs}
    \bibliographystyle{unsrt}
}


\appendix

\setcounter{table}{0}
\setcounter{figure}{0}
\renewcommand{\thetable}{A\arabic{table}}
\renewcommand{\thefigure}{A\arabic{figure}}
\renewcommand{\theHfigure}{AppendixFigure\arabic{figure}}
\renewcommand{\theHtable}{AppendixTable\arabic{table}}

\section{Appendix}
\FloatBarrier

\subsection{Impact Statement}

We introduce the Generalized Induction-head Model (GIM), which improves the performance of fully interpretable models while maintaining transparency, making interpretable models more viable for high-stakes applications. For example, in medical note generation, interpretable models can enhance transparency, enabling clinicians to audit AI-generated text and reduce the risk of hallucinations or biased outputs. Additionally, GIM’s token-level grounding can improve fairness in language models and mitigate bias in automated decision-making. GIM also achieved significant speedups in speculative decoding compared to inference with LLaMA2-70B alone, making it suitable for deployment in compute-limited settings (see \cref{sec:sp}). Despite these advantages, GIM does not fully close the performance gap with black-box models, particularly for tasks requiring extensive reasoning or broad world knowledge. Its reliance on input context may also limit effectiveness in some scenarios where high-quality data is not available.

\subsection{Training of \fuzzyslmlong{}}\label{sec:fuzzyslm_appendix}
\paragraph{Architecture of \fuzzyslmlong{}}
We train two \fuzzyslmlong{}s, one using the GPT-2 tokenizer and the other using the LLaMA-2 tokenizer.
With GPT-2 tokenizer, \fuzzyslmlong{} consists of four transformer layers, whereas it comprises three transformer layers when using LLaMA-2 tokenizer.
Since relative position is crucial for calculating similarity, we incorporate Relative Positional Encoding~\citep{shaw2018self}, with a maximum relative position of 32 for the GPT-2 tokenizer and 64 for the LLaMA-2 tokenizer. The vocabulary embeddings are initialized with those from GPT-2 and LLaMA2-7B, ensuring that the number of heads and embedding dimensions align with the specifications of GPT-2 and LLaMA2-7B.

\paragraph{Creating similarity pair with LLMs} 
For both \fuzzyslmlong{}, we use LLaMA2-7B as a teacher model. OpenWebText and Pile-train\footnote{\scriptsize\url{https://huggingface.co/datasets/monology/pile-uncopyrighted}} datasets for training each \fuzzyslmlong{} that use GPT-2 or LLaMA-2 tokenizer.
During training, we randomly sample sequences of 32 or 64 tokens with batch size of 128 or 256, resulting in 4,096 or 16,384 next-token prediction probabilities per batch. From these, we sample distant 3,584 or 4,096 queries and 512 keys and create similarity pairs ($3,584 \times 512$ or $4,096 \times 512$) by calculating similarity based on \Cref{eq:main}.
The models are trained using a combination of CE loss and reverse KLD loss, with equal weights (1.0). We adopt most of the training settings from the codebase\footnote{\scriptsize\url{https://github.com/karpathy/minGPT}} for training. Gradients are accumulated over 16 iterations, and we use the AdamW optimizer~\citep{loshchilov2018decoupled} with a learning rate of 0.0001 and a weight decay of 0.1. The learning rate follows a cosine schedule with a warmup over the first 1,000 iterations, and training continues for 15,000 or 20,000 iterations. Training is conducted on four NVIDIA A100 GPUs.

\new{
\paragraph{Ablation study on \fuzzyslmlong{} training}
We conduct an ablation study on the positional encoding strategy and training process of \fuzzyslmlong{} using the OpenWebText dataset to distill it from LLaMA-2-7B. The study evaluates the contributions of Relative Positional Encoding, reverse KLD loss, and CE loss to the model’s effectiveness.
As shown in \Cref{tab:ablation_fmm}, next-token prediction accuracy improves significantly when both reverse KLD and CE losses are included, demonstrating their complementary roles in optimizing the \fuzzyslmlong{}. With CE loss, Forward KLD loss is less effective than reverse KLD loss. Furthermore, using Relative Positional Encoding instead of Sinusoidal Positional Encoding leads to better performance, highlighting the advantages of incorporating relative positional information for enhanced fuzzy matching capabilities.
}

We also perform an ablation study on the training data. \Cref{tab:ablation_fmm_data} shows that the performance difference between OpenWebText and Pile-train is minimal. When trained on Pile-of-Law~\citep{henderson2022pile}, a more domain-specific corpus, \fuzzyslmlong{} exhibits slightly lower performance. This suggests that domain specificity may slightly limit the generalization ability of the fuzzy matching module. Nevertheless, the approach remains robust even with more domain-specific training data.

\begin{table}[t]
\caption{{Ablation study on training of \fuzzyslmlong{}. Next-token accuracy (\%) of \fuzzyinduction{} on the BabyLM-test is reported. LLaMA-2 tokenizer is used.}}
\label{tab:ablation_fmm}
\small
\begin{center}
\begin{tabular}{cccccc}\toprule
Positional Encoding &Reverse KLD loss & Forward KLD loss &CE loss &Accuracy \\\midrule
Relative &\checkmark &&\checkmark &43.2 \\
Relative &&\checkmark &\checkmark &42.8 \\
Relative & &&\checkmark &42.7 \\
Relative &\checkmark & &&41.9 \\
Sinusoidal &\checkmark &&\checkmark &37.0 \\
\bottomrule
\end{tabular}
\end{center}
\end{table}

\begin{table}[t]
\caption{{Ablation study on training \fuzzyslmlong{} with different datasets. Next-token accuracy (\%) of \fuzzyinduction{} on the BabyLM-test is reported. LLaMA-2 tokenizer is used.}}
\label{tab:ablation_fmm_data}
\small
\begin{center}
\begin{tabular}{cc}\toprule
Dataset &Accuracy \\\midrule
OpenWebText &43.2 \\
Pile-train &42.7 \\
Pile-of-law &41.8 \\
\bottomrule
\end{tabular}
\end{center}
\end{table}

\subsection{Determination of $\tau$}\label{sec:effective_n_thres}
To build \method{} by integrating the three types of estimations, we first need to determine the threshold for effective $n$, denoted as $\tau$. To identify the optimal value of $\tau$, we conducted cross-validation using the BabyLM training set (100M tokens). BabyLM consists of six datasets: \texttt{open\_subtitles}, \texttt{bnc\_spoken}, \texttt{gutenberg}, \texttt{childes}, \texttt{simple\_wiki}, and \texttt{switchboard}. Since \texttt{switchboard} contains only 2M tokens, we exclude it from the experiment. For the remaining datasets, we use each dataset as a validation set, while the other four are used as the reference corpus to build \infinigram{}. We then compare the performance changes of \infinigram{}, \exactinduction{}, and \fuzzyinduction{} depending on effective $n$. 10k samples are used for evaluating on each dataset.

\begin{figure}[ht]
    \centering
    \includegraphics[width=0.8\linewidth]{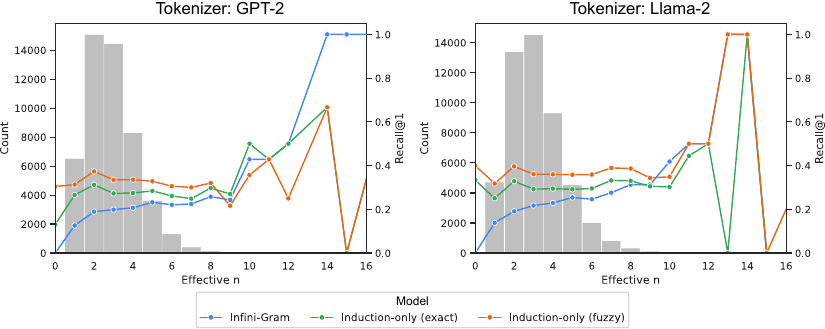}
    \caption{Comparison of next-token accuracy.}
    \label{fig:result_cv}
\end{figure}

As shown in \Cref{fig:result_cv}, \infinigram{} outperforms \exactinduction{} when the effective $n$ exceeds 8 for the GPT-2 tokenizer and 9 for the LLaMA-2 tokenizer. Therefore, we set $\tau$ to 8 and 9 for the respective tokenizers.

\subsection{Language Modeling Results Extended}\label{subsec:lang_model_results_appendix}
\paragraph{Experimental details}
We use diverse datasets as reference corpus for \infinigram{}. We use \infinigram{} that is released by authors\footnote{\scriptsize\url{https://infini-gram.io/api_doc.html}} for Pile-train\footnote{$\texttt{v4\_piletrain\_llama}$} and Pile-val\footnote{$\texttt{v4\_pileval\_llama}$ and $\texttt{v4\_piletrain\_gpt2}$}. For BabyLM-dev and OpenWebText, we build our own \infinigram{}. We use public code to build and inference \infinigram{}\footnote{\scriptsize\url{https://infini-gram.io/pkg_doc.html}} and \exactinduction{}\footnote{\scriptsize\url{https://github.com/AlexWan0/infini-gram/tree/main}}. During inference, the maximum length for exact matching with \infinigram{} is 500, and we use window size $k$ for fuzzy matching as 32 and 64 for GPT-2 and LLaMA-2 tokenizers, respectively.

\new{
\paragraph{Ablation study on \infinigram{} with \method{}}
We conduct an ablation study to assess the impact of each component in \infinigram{} with \method{}. \Cref{tab:ablation_component} reports next-token accuracy when individual components are omitted.
Excluding \fuzzyinduction{} results in a more significant performance drop than removing \exactinduction{}. This underscores the importance of fuzzy matching in handling diverse contexts and improving adaptability, as reflected in \Cref{tab:peformance}, where \fuzzyinduction{} outperforms \exactinduction{}. Since both components act as induction heads, they exhibit complementary roles—when one is removed, the other partially compensates for its absence.
Only when using Pile-train as a reference corpus, omitting \infinigram{} leads to the most substantial performance decline. It is worth noting that when the reference corpus lacks similarity to the test dataset’s distribution (\textit{e.g.}, Pile-val, OpenWebText, and Pile-train), the performance of \infinigram{} falls significantly below the scenario where it is not utilized at all. This highlights the sensitivity of \infinigram{} to the quality and relevance of the reference corpus.
}
\begin{table}[t]
\caption{{Ablation study on components of \textbf{\infinigram{} with \method{}}. Next-token accuracy (\%) on BabyLM-test is reported.}}
\label{tab:ablation_component}
\small
\begin{center}
\begin{tabular}{lcccc}\toprule
Reference Corpus &BabyLM-dev &Pile-val &OpenWebText &Pile-train \\\midrule
\infinigram{} with \method{}&43.1 &42.9 &43.2 &49.4 \\
\midrule
~w/o \fuzzyinduction &42.2 &36.9 &38.3 &46.6 \\
~w/o \exactinduction &43.0 &42.8 &43.1 &49.3 \\
~w/o \infinigram{} &\multicolumn{4}{c}{42.9} \\
\midrule
\infinigram{} &39.0 &19.0 &20.1 &33.5 \\
\bottomrule
\end{tabular}
\end{center}
\end{table}

\subsection{Speculative Decoding}\label{sec:sp}
\setlength{\tabcolsep}{5pt}
\begin{table*}[t]
\caption{Speed of speculative decoding (SP). Accept. denotes the acceptance rate (\%). The mean and standard deviation of 3 runs are reported.}
\small
\label{tab:speculative_decoding}
\begin{center}
\begin{adjustbox}{max width = 1.0\textwidth}
\begin{tabular}{llcccccccc}\toprule
& \multirow{2}{*}{Draft Model} &\multirow{2}{*}{Large Model}  &\multirow{2}{*}{SP} &\multicolumn{3}{c}{BabyLM-test} &\multicolumn{3}{c}{Pile-val} \\\cmidrule{5-10}
&&& &\multirowcell{2}{Accept.\\rate (\%)} & \multicolumn{2}{c}{Speed}&\multirowcell{2}{Accept.\\rate (\%)} & \multicolumn{2}{c}{Speed} \\\cmidrule{6-7}\cmidrule{9-10}
&&& && ms/token ($\downarrow$) &Up ($\uparrow$) && ms/token ($\downarrow$) &Up ($\uparrow$) \\\midrule
\multirow{6}{*}{\rotatebox[origin=c]{90}{A40$\times$1}}&&LLaMA2-7B & &&30.2±0.0 & &&30.2±0.1 & \\
&TinyLLaMA-1.1B &LLaMA2-7B &\checkmark &78.7±0.5 & 21.3±0.0 &1.42 &78.3±0.1 & 21.3±0.6 &1.42 \\
&\fuzzyinduction{} &LLaMA2-7B &\checkmark &74.9±1.1 & 17.7±0.7 &1.71 &71.2±0.5 & 20.1±0.4 &1.50 \\
\cmidrule{2-10}
&&LLaMA2-13B & &&52.4±0.0 & &&52.0±0.2 &\\
&TinyLLaMA-1.1B &LLaMA2-13B &\checkmark &78.2±0.0 & 26.7±0.5 &1.96 &77.6±0.1 & 26.3±0.5 &1.98 \\
&\fuzzyinduction{} &LLaMA2-13B &\checkmark &73.5±0.1 & 24.8±0.1 &2.11 &69.8±0.2 & 27.8±0.1 &1.87 \\
\midrule
\multirow{8}{*}{\rotatebox[origin=c]{90}{H100$\times$2}}&&LLaMA2-13B & &&26.4±0.1 & &&26.3±0.4 & \\
&LLaMA2-7B &LLaMA2-13B &\checkmark &78.9±0.0&24.7±0.0 &1.07 &78.6±0.0&25.1±0.3 &1.05 \\
&TinyLLaMA-1.1B &LLaMA2-13B &\checkmark &78.3±0.1&20.7±0.1& 1.28 &77.6±0.1&21.5±0.1&1.22\\
&\fuzzyinduction{} &LLaMA2-13B &\checkmark &73.2±0.3&13.3±0.2 &1.98 &69.9±0.1&14.9±0.1 &1.77 \\
\cmidrule{2-10}
&&LLaMA2-70B & &&71.2±0.1 & &&71.0±0.2 & \\
&LLaMA2-7B &LLaMA2-70B &\checkmark &77.2±0.2&38.3±0.5 &1.86 &77.8±0.2&37.4±0.3 &1.90 \\
&TinyLLaMA-1.1B &LLaMA2-70B &\checkmark &75.5±0.1&35.3±0.2&2.02&76.3±0.4 & 33.9±0.6 & 2.10\\
&\fuzzyinduction{} &LLaMA2-70B &\checkmark &68.5±0.6&31.4±0.7 &2.27 &66.6±0.6&33.3±0.6 &2.13 \\
\bottomrule
\end{tabular}
\end{adjustbox}
\end{center}
\end{table*}

\method{} offers both interpretability and efficiency. When combined with LLMs in speculative decoding, it enhances prediction accuracy while significantly boosting inference speed.

\paragraph{Experimental details}
To evaluate the efficiency of \fuzzyinduction{}, we compare the inference time for speculative decoding with TinyLLaMA\footnote{\scriptsize \url{https://huggingface.co/TinyLLaMA/TinyLLaMA-1.1B-intermediate-step-1431k-3T}} and LLaMA2-7B~\citep{touvron2023llama2}.
We evaluate speculative decoding by generating up to 1024 tokens, using a prefix of 1024 tokens. The speed of decoding may vary depending on the computational environment. To ensure robust evaluation across different setups, we conduct experiments in two environments: one with a single NVIDIA A40 GPU and 128 CPU cores, and another with two NVIDIA H100 GPUs and 64 CPU cores. Greedy sampling is used for token generation, and each experiment is repeated three times with different random seeds.

\paragraph{Induction improves speculative decoding performance}
\cref{tab:speculative_decoding} demonstrates the speed-up effect of speculative decoding with \fuzzyinduction{}. \fuzzyinduction{} relies solely on the induction power derived from the input context to predict the next token\new{, leading to lower acceptance rates compared to LLMs.} Despite this, its inference speed is remarkably fast, and it often matches the predictions of large models. As a result, the speed improvement can exceed $2\times$ compared to using LLaMA2-70B alone. In most cases, \fuzzyinduction{} achieves even greater speed gains than when using an LLM as a draft model for speculative decoding.

Additionally, we would like to note that speculative decoding with \fuzzyinduction{} and a pretrained LLM not only accelerates the inference speed of the pretrained model but also enables explainable predictions based on the given input context. When accurate predictions can be made through interpretable methods, we utilize this process for interpretability. In more challenging cases, we rely on a larger model that, while less interpretable, delivers better performance for accurate predictions. Thus, this approach provides a balanced method that addresses both interpretability and accuracy, in addition to enhancing efficiency.

\begin{table}[t]
\setlength{\tabcolsep}{4.3pt}
\caption{Speed of speculative decoding (SP). The mean and standard deviation of 3 runs are reported.
} 
\small
\label{tab:speculative_decoding_appendix}
\begin{center}
\begin{adjustbox}{max width = 1.0\textwidth}
\begin{tabular}{lccccccccc}\toprule
\multirow{2}{*}{Draft Model} &\multirow{2}{*}{Large Model} &\multirow{2}{*}{SP}&\multicolumn{2}{c}{BabyLM-test} &\multicolumn{2}{c}{Pile-val} &\multicolumn{2}{c}{FineWeb} \\\cmidrule{4-9}
&&&ms/token ($\downarrow$) &Speed Up ($\uparrow$) &ms/token ($\downarrow$) &Speed Up ($\uparrow$) &ms/token ($\downarrow$) &Speed Up ($\uparrow$) \\\midrule
&LLaMA2-13B & &26.4±0.1 & &26.3±0.4 & \\
\fuzzyinduction{} &LLaMA2-13B &\checkmark &13.3±0.2 &1.98 &14.9±0.1 &1.77 &14.9±0.3 &1.76 \\
\method{} &LLaMA2-13B &\checkmark &23.1±0.4 &1.14 &22.8±0.3 &1.15 &23.0±0.7 &1.14 \\ \midrule
&LLaMA2-70B & &71.2±0.1 & &71.0±0.2 & &71.1±0.2 & \\
\fuzzyinduction{} &LLaMA2-70B &\checkmark &31.4±0.7 &2.27 &33.3±0.6 &2.13 &33.2±1.0 &2.15 \\
\method{} &LLaMA2-70B &\checkmark &42.0±0.7 &1.70 &41.6±1.0 &1.71 &40.4±1.2 &1.76 \\

\bottomrule
\end{tabular}
\end{adjustbox}
\end{center}
\end{table}
\Cref{tab:speculative_decoding_appendix} reports the inference times for \fuzzyinduction{} and \method{} using speculative decoding, with the OpenWebText dataset serving as the reference corpus for \infinigram{}. We find matches with a maximum of 64 tokens for both exact and fuzzy matching. The experiments are conducted on two NVIDIA H100 GPUs and 64 CPU cores.
Although \method{} requires more time for generation on average than \fuzzyinduction{}, it still significantly reduces inference time compared to relying solely on a large model for inference.

\begin{figure}[t]
    \centering
    \includegraphics[width=\textwidth]{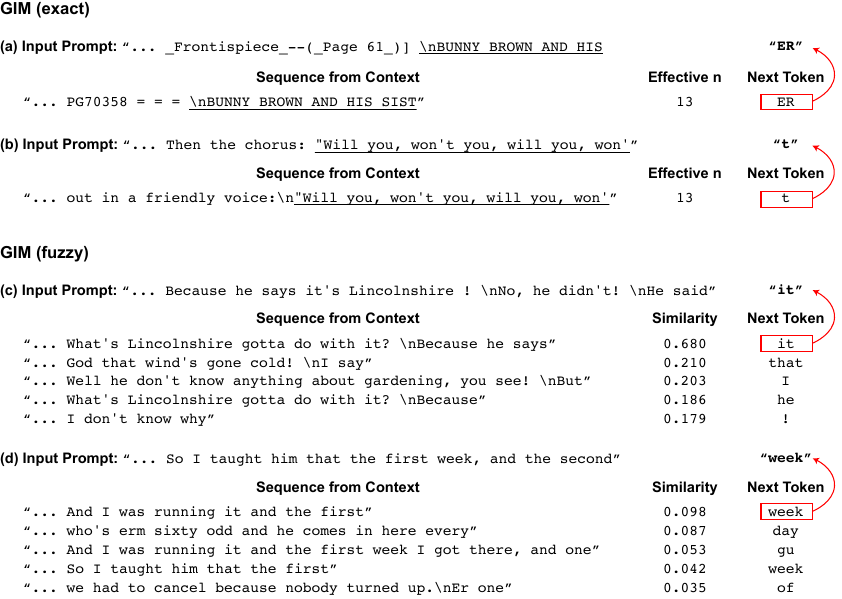}
    \caption{{Examples of explanation of \method{} from BabyLM-test. (a) and (b) show examples of exact matching while (c) and (d) show examples of fuzzy matching. The red box marks the source of the final prediction.}}
    \label{fig:explanation}
\end{figure}

\paragraph{Explanation}
\Cref{fig:explanation} presents several examples of explanations provided by \method{}. Even if an exact match fails to yield a good match, when the probability of subsequent tokens is similar, the fuzzy matching model can predict with high similarity, enabling successful fuzzy matching, enabling successful fuzzy matching, and improving next-token prediction.

\subsection{fMRI results extended}
\label{subsec:fmri_appendix}

\paragraph{Data details}

We analyze publicly available data collected from prior work~\citep{lebel2022natural,tang2023semantic}. Methods from the previous study are summarized here for completeness. Functional magnetic resonance imaging (fMRI) data was recorded from three healthy participants as they listened to English-language podcast stories over Sensimetrics S14 headphones. Participants were only instructed to listen to the stories. No explicit behavioral responses were required.

For the collection of training data, each participant completed approximately 20 hours of listening sessions across 20 separate sessions in which unique stories were presented. This produced 33,000 timepoints per voxel across the entire human cortex. For testing data collection, participants heard two-held-out stories five times each and a third story ten times (one story per session). These repeated measurements were averaged to improve reliability. Signal-to-noise ratios for each voxel were estimated using the mean-explainable variance approach from \citep{nishimoto2017eye}. Analysis was restricted to voxels that were located within 8mm of the cortical mid-surface, yielding about 90,000 voxels per participant.

All participants were healthy adults with normal hearing and gave written informed consent. The study protocol was approved by the Institutional Review Board of the University of Texas at Austin. The scans were acquired on a 3 T Siemens Skyra MRI system at the University of Texas at Austin using a 64-channel Siemens head coil. Functional images were obtained with a gradient-echo EPI sequence (TR = 2.0 s, TE = 30.8 ms, flip angle = 71°, multi-band factor = 2, voxel size = 2.6 mm × 2.6 mm × 2.6 mm, matrix size = 84 × 84, field of view = 220 mm). Anatomical scans were collected using a T1-weighted multi-echo MP-RAGE sequence with 1 mm isotropic voxels, following the standard Freesurfer morphometry protocol~\citep{fischl2012freesurfer}. 

Functional data was preprocessed with FSL 5.0 using the FMRIB Linear Image Registration Tool (FLIRT) for motion correction and alignment. Each participant’s runs were registered to a subject-specific template built from the first functional run of the first session, with all automated registrations manually verified for accuracy. Low-frequency signal drifts were removed using a second-order Savitzky–Golay filter with a 120 s window. To mitigate onset artifacts and detrending issues near scan boundaries, 20 s (10 volumes) were discarded from both the start and end of each run. This eliminated the initial silent period and the first and last 10 s of each story. Each voxel’s mean response was then subtracted, and the remaining signal was normalized to unit variance.

To ensure temporal alignment between linguistic and neural data, word onset times were interpolated to the fMRI sampling rate using Lanczos interpolation with a window size of 3. The hemodynamic response was modeled as a finite impulse response (FIR) with four time lags ($-8$, $-6$, $-4$, and $-2$~s), following the approach of \citep{huth2016natural}. For every subject $x$, voxel $v$, we fit an encoding model $g_{(x,v)}$ to predict the BOLD response $\hat{B}$ from the embedded stimulus, i.e. $\hat{B}_{(x,v)} = g_{(x,v)}(H_i(\mathcal{S}))$. Model evaluation was performed on the held-out test stories, using the trained encoding models to predict and assess voxel responses.

\paragraph{fMRI fuzzy induction head settings}
Similar to the \method{} Matching technique described in~\Cref{subsec:fmri_experimental_setup}, we construct an induction head for fuzzy matching. In the fuzzy setting, we leverage the predicted next-word distributions obtained through fuzzy n-gram matching in the input context ($P_\text{induction}^{\text{(fuzzy)}}$ in~\Cref{eq:cnt}), which we refer to as \textit{Fuzzy Induction Matching}. Specifically, we calculate the cosine similarity between the next-word distributions of the current word and all prior candidate words.

To account for the temporal resolution of fMRI, we apply Lanczos smoothing to the word-level similarity values, aligning these values with the fMRI time scale. This allows us to identify the time point (TR) $t^{*}$ that maximally corresponds to the current time point $t$ based on the highest similarity.

We evaluate several configurations for deriving the next-word distributions, including GPT-2, LLaMa-2, the Fuzzy Matching model with the GPT-2 tokenizer, and the Fuzzy Matching Model with the LLaMA-2 tokenizer. See more details on Fuzzy Matching models in ~\Cref{subsec:in_context_method}.

\paragraph{Extended prediction performance results}

The prediction performance of Fuzzy Induction Matching Models is compared to the performance of the \method{} Matching Models and the Eng1000 baseline in~\Cref{tab:fuzzy_matching}. The Fuzzy Induction Model, in its highest-performing configuration (using the Fuzzy Matching Model with the LLaMa2-70B tokenizer), achieves only a 6.94\% improvement in prediction performance compared to the Eng1000 baseline. The lower relative performance of Fuzzy Induction Matching compared to \method{} Matching may be due to the inherent noise and lower spatial and temporal resolution of fMRI data, which makes it challenging to detect subtle differences in neural activations associated with similar but non-identical stimuli.

\begin{table}[t]
\caption{fMRI Prediction Performance when using fuzzy matching. Error bars show 95\% CI.}
\label{tab:fuzzy_matching}
\begin{center}
\begin{adjustbox}{max width = 1.0\textwidth}
\begin{tabular}{lllcc}\toprule
\multirow{2}{*}{Feature Model} & \multirow{2}{*}{Tokenizer} & \multirow{2}{*}{Matching Model} & \multicolumn{2}{c}{Mean Correlation} \\\cmidrule{4-5}
& & & All Voxels & Top 10\% Voxels \\\midrule
Eng1000         & - & - & $0.072 \pm 0.0004$ & $0.220 \pm 0.0012$ \\
\infinigram{} + Eng1000   & GPT-2 & - & $0.069 \pm 0.0003$ & $0.200 \pm 0.0012$ \\
\method{} Matching + Eng1000       & GPT-2 & - & $0.087 \pm 0.0005$ & $0.265 \pm 0.0011$ \\
\midrule
Fuzzy Induction Matching + Eng1000 & GPT-2 & GPT-2 & $0.076 \pm 0.0004$ & $0.222 \pm 0.0011$ \\
Fuzzy Induction Matching + Eng1000 & LLaMA-2 & LLaMA2-70B & $0.076 \pm 0.0004$ & $0.225 \pm 0.0012$ \\
Fuzzy Induction Matching + Eng1000 & GPT-2 & Fuzzy Matching Model & $0.076 \pm 0.0004$ & $0.216 \pm 0.0011$ \\
Fuzzy Induction Matching + Eng1000 & LLaMA-2 & Fuzzy Matching Model & $0.077 \pm 0.0004$ & $0.223 \pm 0.0012$ \\
\bottomrule
\end{tabular}
\end{adjustbox}
\end{center}
\end{table}

\begin{table}[t]
\centering
\caption{GPT-4 Prompts for Generating and Classifying Categories of Text. Ellipses (...) indicate omitted portions of the full prompts.}
\small
\begin{tabular}{p{5.5cm} p{8cm}}
\toprule
\textbf{Title} & \textbf{Prompt} \\ \midrule
GPT-4 Prompt for Generating Category Descriptions & I have provided two test stories below. Specific phrases from each story have been picked out based on the performance of different encoding models. Can you describe the characteristics of the words and phrases that each category contains? Be specific about the type of words, their context in the story, and any other relevant commonalities. Write succinct descriptions for each category that would allow one to categorize phrases in other such stories accurately. 

Category A: ['sh first she digs into her cutoffs in the', 'both need this right now i', ... ] 

Category B: ['to everything or you make yourself scarce', 'my cigarettes and uh', ...]

Full Story: [['i reached over and secretly'], ['undid my seatbelt'], ...] \\ \midrule

GPT-4 Prompt for Classifying Stages Based on Descriptions & I have attached category descriptions below. Based on the descriptions, in order, go through each short list of words (short phrase) in the story at the end and classify the segments into one of the categories. Rather than listing all the phrases in a category at a time, list each phrase in order and label it as belonging to category A or B.

Category A: Emotionally, or Narratively Critical ...

Category B: Brief, Stand-Alone Phrases ...

Full Story: [['i reached over and secretly'], ['undid my seatbelt'], ...] \\ \bottomrule
\end{tabular}
\end{table}

\newpage

\begin{figure}[p]
    \centering
    \includegraphics[width=15cm,height=25cm,keepaspectratio]{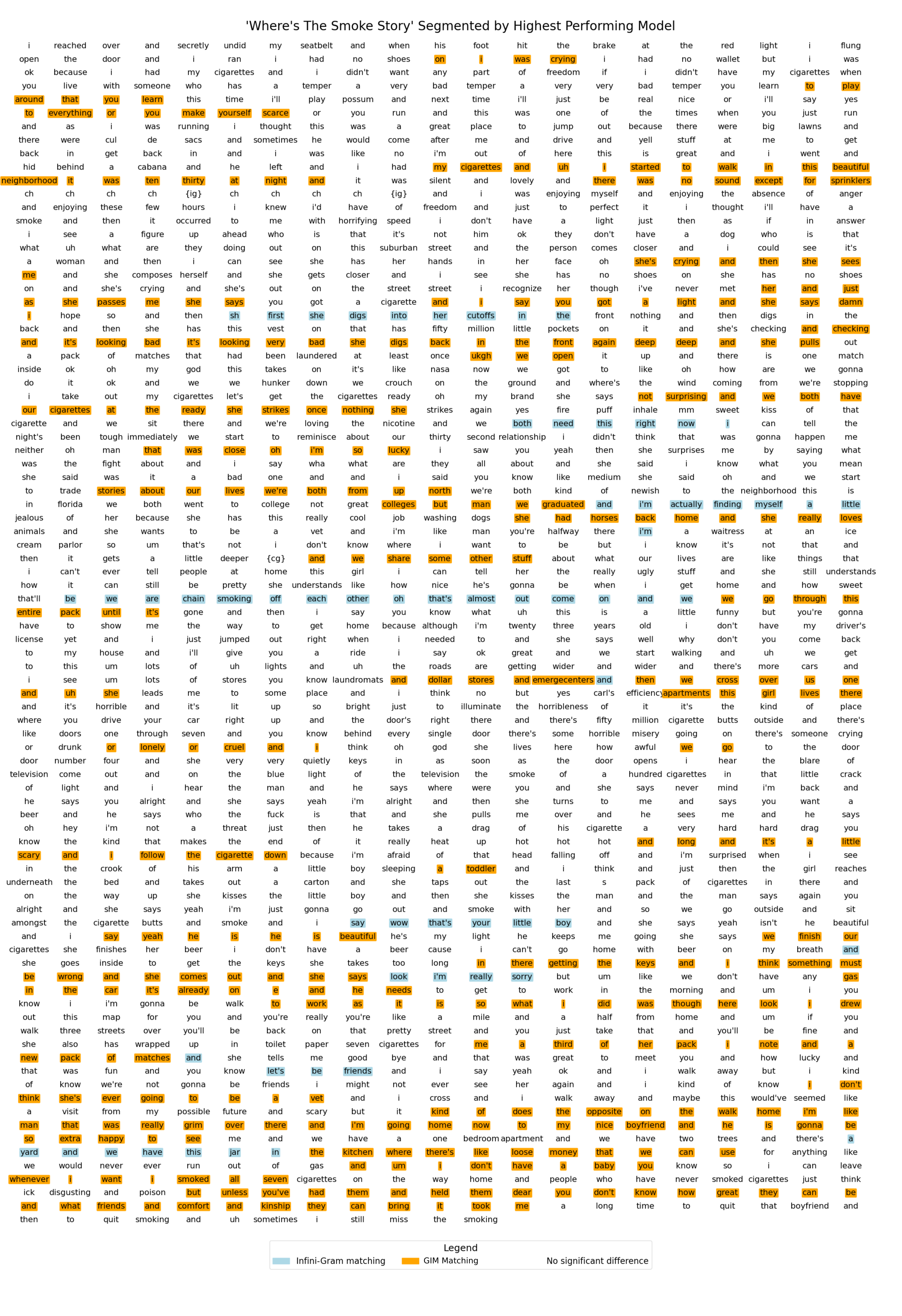}
    \caption{Test story 1 (\textit{Where's There's Smoke}), highlighted in regions where the Infini-Gram matching and \method{} matching models exceed baseline performance, measured by the average absolute error across voxels, by more than one standard deviation.}
    \label{fig:full_story_seg}
\end{figure}

\begin{figure}[p]
    \centering
    \includegraphics[width=14cm,height=24cm,keepaspectratio]{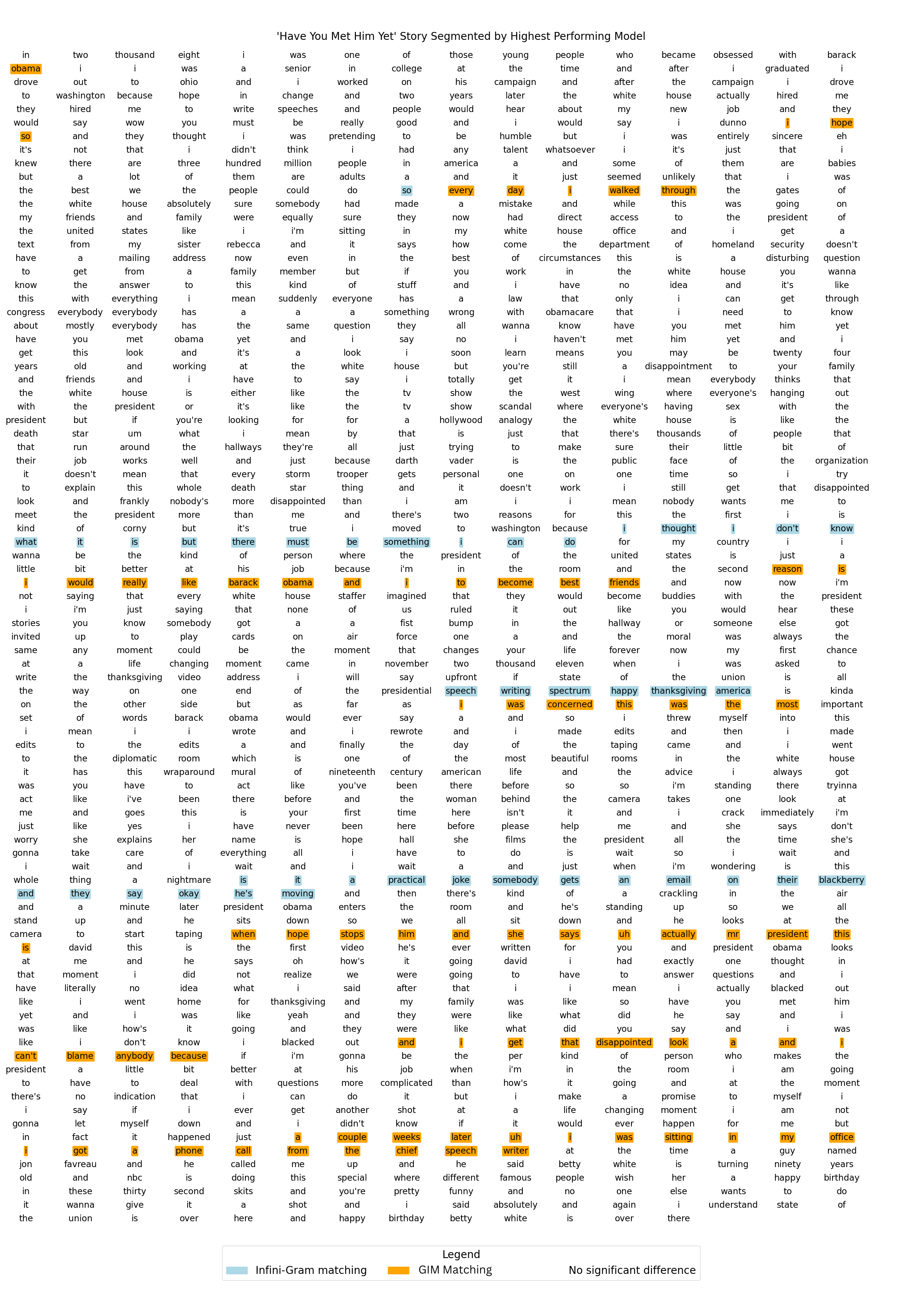}
    \caption{The first section of test story 2 (\textit{Have You Met Him Yet}), highlighted in regions where the Infini-Gram and \method{} matching models exceed baseline performance.}
    \label{fig:full_story_seg_a}
\end{figure}

\begin{figure}[p]
    \centering
    \includegraphics[width=14cm,height=24cm,keepaspectratio]{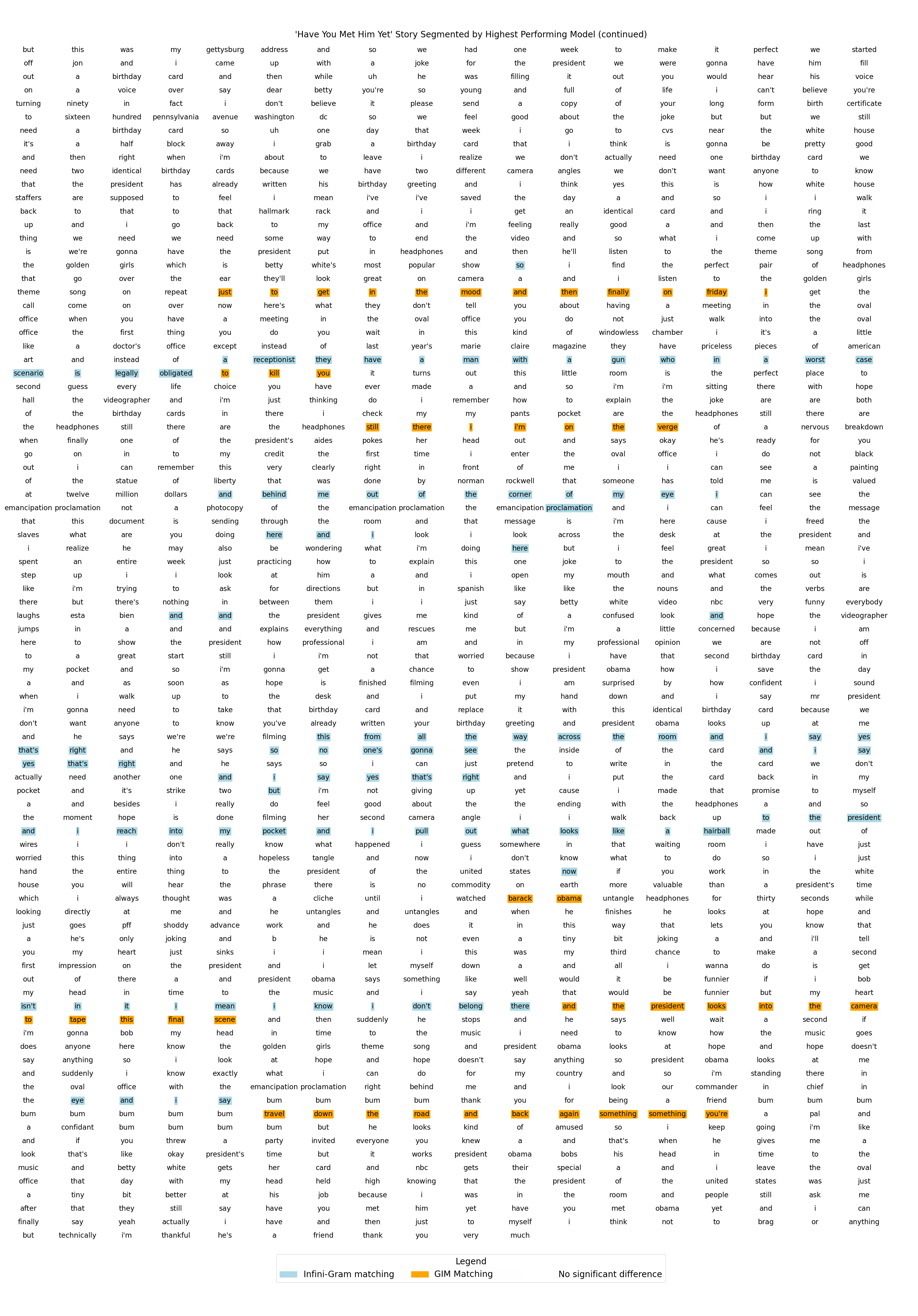}
    \caption{The second section of test story 2 (\textit{Have You Met Him Yet}), highlighted in regions where the Infini-Gram and \method{} matching models exceed baseline performance.}
    \label{fig:full_story_seg_b}
\end{figure}


\newpage
\clearpage
\section*{NeurIPS Paper Checklist}

\begin{enumerate}

\item {\bf Claims}
    \item[] Question: Do the main claims made in the abstract and introduction accurately reflect the paper's contributions and scope?
    \item[] Answer: \answerYes{} 
    \item[] Justification: We reviewed the abstract and introduction to ensure they accurately reflect the paper's contributions and scope.
    \item[] Guidelines:
    \begin{itemize}
        \item The answer NA means that the abstract and introduction do not include the claims made in the paper.
        \item The abstract and/or introduction should clearly state the claims made, including the contributions made in the paper and important assumptions and limitations. A No or NA answer to this question will not be perceived well by the reviewers. 
        \item The claims made should match theoretical and experimental results, and reflect how much the results can be expected to generalize to other settings. 
        \item It is fine to include aspirational goals as motivation as long as it is clear that these goals are not attained by the paper. 
    \end{itemize}

\item {\bf Limitations}
    \item[] Question: Does the paper discuss the limitations of the work performed by the authors?
    \item[] Answer: \answerYes{} 
    \item[] Justification: We discussed the limitations of our work in the Discussion section.
    \item[] Guidelines:
    \begin{itemize}
        \item The answer NA means that the paper has no limitation while the answer No means that the paper has limitations, but those are not discussed in the paper. 
        \item The authors are encouraged to create a separate "Limitations" section in their paper.
        \item The paper should point out any strong assumptions and how robust the results are to violations of these assumptions (e.g., independence assumptions, noiseless settings, model well-specification, asymptotic approximations only holding locally). The authors should reflect on how these assumptions might be violated in practice and what the implications would be.
        \item The authors should reflect on the scope of the claims made, e.g., if the approach was only tested on a few datasets or with a few runs. In general, empirical results often depend on implicit assumptions, which should be articulated.
        \item The authors should reflect on the factors that influence the performance of the approach. For example, a facial recognition algorithm may perform poorly when image resolution is low or images are taken in low lighting. Or a speech-to-text system might not be used reliably to provide closed captions for online lectures because it fails to handle technical jargon.
        \item The authors should discuss the computational efficiency of the proposed algorithms and how they scale with dataset size.
        \item If applicable, the authors should discuss possible limitations of their approach to address problems of privacy and fairness.
        \item While the authors might fear that complete honesty about limitations might be used by reviewers as grounds for rejection, a worse outcome might be that reviewers discover limitations that aren't acknowledged in the paper. The authors should use their best judgment and recognize that individual actions in favor of transparency play an important role in developing norms that preserve the integrity of the community. Reviewers will be specifically instructed to not penalize honesty concerning limitations.
    \end{itemize}

\item {\bf Theory assumptions and proofs}
    \item[] Question: For each theoretical result, does the paper provide the full set of assumptions and a complete (and correct) proof?
    \item[] Answer: \answerNA{} 
    \item[] Justification: Our paper does not include theoretical results.
    \item[] Guidelines:
    \begin{itemize}
        \item The answer NA means that the paper does not include theoretical results. 
        \item All the theorems, formulas, and proofs in the paper should be numbered and cross-referenced.
        \item All assumptions should be clearly stated or referenced in the statement of any theorems.
        \item The proofs can either appear in the main paper or the supplemental material, but if they appear in the supplemental material, the authors are encouraged to provide a short proof sketch to provide intuition. 
        \item Inversely, any informal proof provided in the core of the paper should be complemented by formal proofs provided in appendix or supplemental material.
        \item Theorems and Lemmas that the proof relies upon should be properly referenced. 
    \end{itemize}

    \item {\bf Experimental result reproducibility}
    \item[] Question: Does the paper fully disclose all the information needed to reproduce the main experimental results of the paper to the extent that it affects the main claims and/or conclusions of the paper (regardless of whether the code and data are provided or not)?
    \item[] Answer: \answerYes{} 
    \item[] Justification: We included all necessary information to reproduce the experimental results in the main text and the appendix.
    \item[] Guidelines:
    \begin{itemize}
        \item The answer NA means that the paper does not include experiments.
        \item If the paper includes experiments, a No answer to this question will not be perceived well by the reviewers: Making the paper reproducible is important, regardless of whether the code and data are provided or not.
        \item If the contribution is a dataset and/or model, the authors should describe the steps taken to make their results reproducible or verifiable. 
        \item Depending on the contribution, reproducibility can be accomplished in various ways. For example, if the contribution is a novel architecture, describing the architecture fully might suffice, or if the contribution is a specific model and empirical evaluation, it may be necessary to either make it possible for others to replicate the model with the same dataset, or provide access to the model. In general. releasing code and data is often one good way to accomplish this, but reproducibility can also be provided via detailed instructions for how to replicate the results, access to a hosted model (e.g., in the case of a large language model), releasing of a model checkpoint, or other means that are appropriate to the research performed.
        \item While NeurIPS does not require releasing code, the conference does require all submissions to provide some reasonable avenue for reproducibility, which may depend on the nature of the contribution. For example
        \begin{enumerate}
            \item If the contribution is primarily a new algorithm, the paper should make it clear how to reproduce that algorithm.
            \item If the contribution is primarily a new model architecture, the paper should describe the architecture clearly and fully.
            \item If the contribution is a new model (e.g., a large language model), then there should either be a way to access this model for reproducing the results or a way to reproduce the model (e.g., with an open-source dataset or instructions for how to construct the dataset).
            \item We recognize that reproducibility may be tricky in some cases, in which case authors are welcome to describe the particular way they provide for reproducibility. In the case of closed-source models, it may be that access to the model is limited in some way (e.g., to registered users), but it should be possible for other researchers to have some path to reproducing or verifying the results.
        \end{enumerate}
    \end{itemize}

\item {\bf Open access to data and code}
    \item[] Question: Does the paper provide open access to the data and code, with sufficient instructions to faithfully reproduce the main experimental results, as described in supplemental material?
    \item[] Answer: \answerYes{} 
    \item[] Justification: We included the URL to the publicly available code in the main text. Additionally, experimental details are demonstrated in the main text and the appendix.
    \item[] Guidelines:
    \begin{itemize}
        \item The answer NA means that paper does not include experiments requiring code.
        \item Please see the NeurIPS code and data submission guidelines (\url{https://nips.cc/public/guides/CodeSubmissionPolicy}) for more details.
        \item While we encourage the release of code and data, we understand that this might not be possible, so “No” is an acceptable answer. Papers cannot be rejected simply for not including code, unless this is central to the contribution (e.g., for a new open-source benchmark).
        \item The instructions should contain the exact command and environment needed to run to reproduce the results. See the NeurIPS code and data submission guidelines (\url{https://nips.cc/public/guides/CodeSubmissionPolicy}) for more details.
        \item The authors should provide instructions on data access and preparation, including how to access the raw data, preprocessed data, intermediate data, and generated data, etc.
        \item The authors should provide scripts to reproduce all experimental results for the new proposed method and baselines. If only a subset of experiments are reproducible, they should state which ones are omitted from the script and why.
        \item At submission time, to preserve anonymity, the authors should release anonymized versions (if applicable).
        \item Providing as much information as possible in supplemental material (appended to the paper) is recommended, but including URLs to data and code is permitted.
    \end{itemize}

\item {\bf Experimental setting/details}
    \item[] Question: Does the paper specify all the training and test details (e.g., data splits, hyperparameters, how they were chosen, type of optimizer, etc.) necessary to understand the results?
    \item[] Answer: \answerYes{} 
    \item[] Justification: We specified all details in the appendix.
    \item[] Guidelines:
    \begin{itemize}
        \item The answer NA means that the paper does not include experiments.
        \item The experimental setting should be presented in the core of the paper to a level of detail that is necessary to appreciate the results and make sense of them.
        \item The full details can be provided either with the code, in appendix, or as supplemental material.
    \end{itemize}

\item {\bf Experiment statistical significance}
    \item[] Question: Does the paper report error bars suitably and correctly defined or other appropriate information about the statistical significance of the experiments?
    \item[] Answer: \answerYes{} 
    \item[] Justification: We added error bars in plots or reported standard deviations in tables.
    \item[] Guidelines:
    \begin{itemize}
        \item The answer NA means that the paper does not include experiments.
        \item The authors should answer "Yes" if the results are accompanied by error bars, confidence intervals, or statistical significance tests, at least for the experiments that support the main claims of the paper.
        \item The factors of variability that the error bars are capturing should be clearly stated (for example, train/test split, initialization, random drawing of some parameter, or overall run with given experimental conditions).
        \item The method for calculating the error bars should be explained (closed form formula, call to a library function, bootstrap, etc.)
        \item The assumptions made should be given (e.g., Normally distributed errors).
        \item It should be clear whether the error bar is the standard deviation or the standard error of the mean.
        \item It is OK to report 1-sigma error bars, but one should state it. The authors should preferably report a 2-sigma error bar than state that they have a 96\% CI, if the hypothesis of Normality of errors is not verified.
        \item For asymmetric distributions, the authors should be careful not to show in tables or figures symmetric error bars that would yield results that are out of range (e.g. negative error rates).
        \item If error bars are reported in tables or plots, The authors should explain in the text how they were calculated and reference the corresponding figures or tables in the text.
    \end{itemize}

\item {\bf Experiments compute resources}
    \item[] Question: For each experiment, does the paper provide sufficient information on the computer resources (type of compute workers, memory, time of execution) needed to reproduce the experiments?
    \item[] Answer: \answerYes{} 
    \item[] Justification: We included the information on the computer resources in the appendix.
    \item[] Guidelines:
    \begin{itemize}
        \item The answer NA means that the paper does not include experiments.
        \item The paper should indicate the type of compute workers CPU or GPU, internal cluster, or cloud provider, including relevant memory and storage.
        \item The paper should provide the amount of compute required for each of the individual experimental runs as well as estimate the total compute. 
        \item The paper should disclose whether the full research project required more compute than the experiments reported in the paper (e.g., preliminary or failed experiments that didn't make it into the paper). 
    \end{itemize}
    
\item {\bf Code of ethics}
    \item[] Question: Does the research conducted in the paper conform, in every respect, with the NeurIPS Code of Ethics \url{https://neurips.cc/public/EthicsGuidelines}?
    \item[] Answer: \answerYes{} 
    \item[] Justification: We have read through the Code of Ethics and confirmed that our research conforms to them.
    \item[] Guidelines:
    \begin{itemize}
        \item The answer NA means that the authors have not reviewed the NeurIPS Code of Ethics.
        \item If the authors answer No, they should explain the special circumstances that require a deviation from the Code of Ethics.
        \item The authors should make sure to preserve anonymity (e.g., if there is a special consideration due to laws or regulations in their jurisdiction).
    \end{itemize}

\item {\bf Broader impacts}
    \item[] Question: Does the paper discuss both potential positive societal impacts and negative societal impacts of the work performed?
    \item[] Answer: \answerYes{} 
    \item[] Justification: We discussed broader impacts in the Appendix and limitations in the Discussion section.
    \item[] Guidelines:
    \begin{itemize}
        \item The answer NA means that there is no societal impact of the work performed.
        \item If the authors answer NA or No, they should explain why their work has no societal impact or why the paper does not address societal impact.
        \item Examples of negative societal impacts include potential malicious or unintended uses (e.g., disinformation, generating fake profiles, surveillance), fairness considerations (e.g., deployment of technologies that could make decisions that unfairly impact specific groups), privacy considerations, and security considerations.
        \item The conference expects that many papers will be foundational research and not tied to particular applications, let alone deployments. However, if there is a direct path to any negative applications, the authors should point it out. For example, it is legitimate to point out that an improvement in the quality of generative models could be used to generate deepfakes for disinformation. On the other hand, it is not needed to point out that a generic algorithm for optimizing neural networks could enable people to train models that generate Deepfakes faster.
        \item The authors should consider possible harms that could arise when the technology is being used as intended and functioning correctly, harms that could arise when the technology is being used as intended but gives incorrect results, and harms following from (intentional or unintentional) misuse of the technology.
        \item If there are negative societal impacts, the authors could also discuss possible mitigation strategies (e.g., gated release of models, providing defenses in addition to attacks, mechanisms for monitoring misuse, mechanisms to monitor how a system learns from feedback over time, improving the efficiency and accessibility of ML).
    \end{itemize}
    
\item {\bf Safeguards}
    \item[] Question: Does the paper describe safeguards that have been put in place for responsible release of data or models that have a high risk for misuse (e.g., pretrained language models, image generators, or scraped datasets)?
    \item[] Answer: \answerNA{} 
    \item[] Justification: We did not release any data or models.
    \item[] Guidelines:
    \begin{itemize}
        \item The answer NA means that the paper poses no such risks.
        \item Released models that have a high risk for misuse or dual-use should be released with necessary safeguards to allow for controlled use of the model, for example by requiring that users adhere to usage guidelines or restrictions to access the model or implementing safety filters. 
        \item Datasets that have been scraped from the Internet could pose safety risks. The authors should describe how they avoided releasing unsafe images.
        \item We recognize that providing effective safeguards is challenging, and many papers do not require this, but we encourage authors to take this into account and make a best faith effort.
    \end{itemize}

\item {\bf Licenses for existing assets}
    \item[] Question: Are the creators or original owners of assets (e.g., code, data, models), used in the paper, properly credited and are the license and terms of use explicitly mentioned and properly respected?
    \item[] Answer: \answerYes{} 
    \item[] Justification: We included citations for all cited papers.
    \item[] Guidelines:
    \begin{itemize}
        \item The answer NA means that the paper does not use existing assets.
        \item The authors should cite the original paper that produced the code package or dataset.
        \item The authors should state which version of the asset is used and, if possible, include a URL.
        \item The name of the license (e.g., CC-BY 4.0) should be included for each asset.
        \item For scraped data from a particular source (e.g., website), the copyright and terms of service of that source should be provided.
        \item If assets are released, the license, copyright information, and terms of use in the package should be provided. For popular datasets, \url{paperswithcode.com/datasets} has curated licenses for some datasets. Their licensing guide can help determine the license of a dataset.
        \item For existing datasets that are re-packaged, both the original license and the license of the derived asset (if it has changed) should be provided.
        \item If this information is not available online, the authors are encouraged to reach out to the asset's creators.
    \end{itemize}

\item {\bf New assets}
    \item[] Question: Are new assets introduced in the paper well documented and is the documentation provided alongside the assets?
    \item[] Answer: \answerNA{} 
    \item[] Justification: We did not introduce any new assets in the paper.
    \item[] Guidelines:
    \begin{itemize}
        \item The answer NA means that the paper does not release new assets.
        \item Researchers should communicate the details of the dataset/code/model as part of their submissions via structured templates. This includes details about training, license, limitations, etc. 
        \item The paper should discuss whether and how consent was obtained from people whose asset is used.
        \item At submission time, remember to anonymize your assets (if applicable). You can either create an anonymized URL or include an anonymized zip file.
    \end{itemize}

\item {\bf Crowdsourcing and research with human subjects}
    \item[] Question: For crowdsourcing experiments and research with human subjects, does the paper include the full text of instructions given to participants and screenshots, if applicable, as well as details about compensation (if any)? 
    \item[] Answer: \answerNA{} 
    \item[] Justification: This paper does not include such experiments or researches.
    \item[] Guidelines:
    \begin{itemize}
        \item The answer NA means that the paper does not involve crowdsourcing nor research with human subjects.
        \item Including this information in the supplemental material is fine, but if the main contribution of the paper involves human subjects, then as much detail as possible should be included in the main paper. 
        \item According to the NeurIPS Code of Ethics, workers involved in data collection, curation, or other labor should be paid at least the minimum wage in the country of the data collector. 
    \end{itemize}

\item {\bf Institutional review board (IRB) approvals or equivalent for research with human subjects}
    \item[] Question: Does the paper describe potential risks incurred by study participants, whether such risks were disclosed to the subjects, and whether Institutional Review Board (IRB) approvals (or an equivalent approval/review based on the requirements of your country or institution) were obtained?
    \item[] Answer: \answerNA{} 
    \item[] Justification: This paper does not encompass such potential risks.
    \item[] Guidelines:
    \begin{itemize}
        \item The answer NA means that the paper does not involve crowdsourcing nor research with human subjects.
        \item Depending on the country in which research is conducted, IRB approval (or equivalent) may be required for any human subjects research. If you obtained IRB approval, you should clearly state this in the paper. 
        \item We recognize that the procedures for this may vary significantly between institutions and locations, and we expect authors to adhere to the NeurIPS Code of Ethics and the guidelines for their institution. 
        \item For initial submissions, do not include any information that would break anonymity (if applicable), such as the institution conducting the review.
    \end{itemize}

\item {\bf Declaration of LLM usage}
    \item[] Question: Does the paper describe the usage of LLMs if it is an important, original, or non-standard component of the core methods in this research? Note that if the LLM is used only for writing, editing, or formatting purposes and does not impact the core methodology, scientific rigorousness, or originality of the research, declaration is not required.
    \item[] Answer: \answerNA{} 
    \item[] Justification: LLMs were not employed for purposes integral to the central aspects of this research.
    \item[] Guidelines:
    \begin{itemize}
        \item The answer NA means that the core method development in this research does not involve LLMs as any important, original, or non-standard components.
        \item Please refer to our LLM policy (\url{https://neurips.cc/Conferences/2025/LLM}) for what should or should not be described.
    \end{itemize}

\end{enumerate}

\end{document}